\documentclass{article}

     \usepackage[final]{neurips_2019}

\usepackage[utf8]{inputenc} 
\usepackage[cmex10]{amsmath}
\usepackage[T1]{fontenc}    
\usepackage{hyperref}       
\usepackage{booktabs}       
\usepackage{amssymb,amsfonts,bbm}       
\usepackage{mathabx}
\usepackage{graphicx}

\newcommand{\Hyp}{\mathsf{C}}
\newcommand{\Exp}{\mathsf{E}}
\newcommand{\Pro}{\mathsf{P}}
\newcommand{\D}{\mathsf{D}}
\newcommand{\f}{\mathsf{f}}
\newcommand{\p}{\mathsf{p}}
\renewcommand{\d}{\mathsf{d}}

\newcommand{\ind}[1]{\mathbbm{1}_{\{#1\}}}     

\DeclareFontFamily{OT1}{pzc}{}
\DeclareFontShape{OT1}{pzc}{m}{it}{<-> s * [1.200] pzcmi7t}{}
\DeclareMathAlphabet{\mathpzc}{OT1}{pzc}{m}{it}

\newcommand{\pzA}{{A}}
\newcommand{\pa}{{a}}
\newcommand{\pB}{{B}}
\newcommand{\pb}{{b}}
\newcommand{\pG}{\mathpzc{B}}
\newcommand{\pg}{\mathpzc{b}}
\newcommand{\pH}{\mathpzc{A}}
\newcommand{\ph}{\mathpzc{a}}
\newcommand{\pM}{\mathpzc{M}}
\newcommand{\pzm}{\mathpzc{m}}
\newcommand{\pN}{\mathpzc{N}}
\newcommand{\pn}{\mathpzc{n}}

\title{Optimizing Shallow Networks for\\ Binary Classification}

\author{%
  Kalliopi~Basioti ~\&~ George V. Moustakides\\
  Department of Computer Science\\
  Rutgers University\\
  Piscataway, NJ 08854, USA \\
  \texttt{(kib21@scarletmail, gm463).rutgers.edu} \\
}

\begin{document}

\maketitle

\begin{abstract}
Data driven classification that relies on neural networks is based on optimization criteria that involve some form of distance between the output of the network and the desired label. Using the same mathematical analysis, for a multitude of such measures, we can show that their optimum solution matches the ideal likelihood ratio test classifier. In this work we introduce a different family of optimization problems which is not covered by the existing approaches and, therefore, opens possibilities for new training algorithms for neural network based classification. We give examples that lead to algorithms that are simple in implementation, exhibit stable convergence characteristics and are antagonistic to the most popular existing techniques.
\end{abstract}


\section{Binary classification as hypothesis testing}
\vskip-0.3cm 
Classification, when there is available a complete probabilistic description of the classes of interest, coincides with statistical hypothesis testing. For this reason we revisit the basic results of this simple and interesting theory and use them as basis in order to build an equivalent data driven version of the problem.

Assume we observe a random vector $X$ for which we distinguish two possible scenarios (classes or hypotheses) $\Hyp_1,\Hyp_2$ regarding its probabilistic description
\vskip-0.1cm\noindent
\hskip0.5cm$\Hyp_1:~~X\sim\f_1(X),~\p_1$\\
\null\hskip0.5cm$\Hyp_2:~~X\sim\f_2(X),~\p_2$.
\vskip-0.1cm\noindent
Functions $\f_1(X),\f_2(X)$ are the probability densities that capture the statistical behavior of $X$ under the corresponding classes and $\p_1,\p_2$ with $\p_1+\p_2=1$ express the prior probability of occurrence of each class.

We are interested in developing a mechanism which, every time we observe a vector $X$, will assign the label 1 or 2 to $X$ in an effort to identify the class the vector is coming from. Specifically, we are looking for a function $\d(X)$ with values in the set $\{1,2\}$ and we would like to select it properly. Following a Bayesian approach, each function$\d(X)$ produces labeling errors with the corresponding error probability being equal to
\begin{equation*}
\Pro_{\rm err}(\d)=\p_1\Pro_1\big(\d(X)=2\big)+\p_2\Pro_2\big(\d(X)=1\big)=
\p_1\Exp_1[\ind{\d(X)=2}]+\p_2\Exp_2[\ind{\d(X)=1}],
\end{equation*}
where $\Pro_k(\cdot),k=1,2$ denotes probability under the density $\f_k(X)$, $\Exp_k[\cdot]$ the corresponding expectation and $\mathbbm{1}_A$ the indicator of the event $A$. 

An optimum classifier is obtained if we select $\d(X)$ to \textit{minimize the error probability $\Pro_{\rm err}(\d)$}. It is known [11, Pages 26--28] that the optimum $\d(X)$ is the well celebrated likelihood ratio test (LRT)
\begin{equation}
\d_o(X)=\left\{\begin{array}{cl}
1,&\text{if}~\frac{\f_1(X)}{\f_2(X)}\geq\frac{\p_2}{\p1}\\
2,&\text{if}~\frac{\f_1(X)}{\f_2(X)}<\frac{\p_2}{\p1},
\end{array}\right.~\text{equivalently}~
\d_o(X)=\left\{\begin{array}{cl}
1,&\text{if}~\p_1\f_1(X)-\p_2\f_2(X)\geq0\\
2,&\text{if}~\p_1\f_1(X)-\p_2\f_2(X)<0,
\end{array}\right.
\label{eq:0}
\end{equation}
As we can see, optimum classification is achieved by consulting the \textit{sign} of $\p_1\f_1(X)-\p_2\f_2(X)$. From \eqref{eq:0} we understand that we could formulate the classification problem as 
\begin{equation}
\d(X)=\left\{\begin{array}{cl}
1,&\text{if}~\D(X)\geq0\\
2,&\text{if}~\D(X)<0,
\end{array}\right.~~\Pro_{\rm err}(\D)=\p_1\Exp_1[\ind{\D(X)<0}]+\p_2\Exp_2[\ind{\D(X)\geq0}]
\label{eq:class}
\end{equation}
where $\D(X)$ is some scalar function and look for the optimum $\D(X)$. Clearly, restricting ourselves to this smaller class does not inflict any performance loss since the optimum classifier (LRT), as pointed out in \eqref{eq:0}, can be put exactly under this form with $\D_o(X)=\p_1\f_1(X)-\p_2\f_2(X)$.

\subsection{Alternative optimization problems}
\vskip-0.2cm To find the optimum $\D(X)$ one would work, directly, with the error probability $\Pro_{\rm err}(\D)$ and attempt to minimize it. This, obviously, will lead to $\D_o(X)=\p_1\f_1(X)-\p_2\f_2(X)$ when the probability densities and the priors are known. When, however, this information is not to our disposal and we are in the pure data driven case, the same optimization problem is completely unsuitable. The reason is that, in order to solve it, we are going to limit $\D(X)$ to some parametric family of functions (as neural networks) and employ (stochastic) \textit{gradient}-type algorithms to optimize for the parameters. This would inevitably require the computation of gradients of indicator functions. Unfortunately, the latter are notorious for having gradients that cannot be used in numerical computations since the gradients are either 0 or at border points, due to discontinuity, they do not exist.

It is because of the above reason that the minimization of the error probability is abandoned in favor of different optimization problem where gradients are well defined. Of course there is an important property that needs to be satisfied by any alternative approach:
\vskip\parskip
\centerline{\textit{The solution of the alternative optimization must be equivalent to LRT}} 
\centerline{\textit{ in order to be useful for the classification problem.}}
Otherwise it will produce ``suboptimum'' classification results. Fortunately, there exists a significant number of optimization problems proposed in the literature that satisfy this basic requirement. Actually, the main scope of this work is to offer additional classes of optimization problems that enjoy the same basic property and can therefore be used to design neural networks.

Regarding existing alternative optimizations, in [1], [10] one can find an interesting mathematical analysis that treats cases that can be put under the following form
\begin{equation}
\mathcal{J}(\D)=\p_1\Exp_1[\phi\big(\D(X)\big)]+\p_2\Exp_2[\phi\big(-\D(X)\big)],~~\min_{\D}\mathcal{J}(\D),
\label{eq:old}
\end{equation}
where $\phi(z)$ is a scalar function. Not every $\phi(z)$ satisfies the basic requirement that the solution $\D_o(X)$ of the minimization in \eqref{eq:old} is equivalent to LRT. 
\begin{table}[h!]\small
\caption{Nonlinearities and optimum classifier functions}
\label{tab:0}
\vskip-0.2cm
\centering
\begin{tabular}{lll}
\toprule
\multicolumn{1}{c}{$\phi(z)$}&\multicolumn{1}{c}{$\D_o(X)$}&References\\
\midrule
$|1-z|$&${\rm sign}(\p_1\f_1-\p_2f_2)$&[5]\\
\midrule
$|1-z|^\rho,\rho>1$&$\frac{(\p_1\f_1)^{\frac{1}{\rho-1}}-(\p_2f_2)^{\frac{1}{\rho-1}}}{(\p_1\f_1)^{\frac{1}{\rho-1}}+(\p_2f_2)^{{\frac{1}{\rho-1}}}}$&[6] (for $\rho=2$)\\
\midrule
$(1-z)^+$&${\rm sign}(\p_1\f_1-\p_2f_2)$&[12]\\
\midrule
$\big((1-z)^+\big)^\rho,\rho>1$&$\frac{(\p_1\f_1)^{\frac{1}{\rho-1}}-(\p_2f_2)^{\frac{1}{\rho-1}}}{(\p_1\f_1)^{\frac{1}{\rho-1}}+(\p_2f_2)^{{\frac{1}{\rho-1}}}}$&[4]\\
\bottomrule
\end{tabular}
\end{table}
Table\,\ref{tab:0} depicts characteristic examples of $\phi(z)$ and the corresponding optimum $\D_o(X)$ where this desirable property is indeed valid. As we can verify, it is always possible from $\D_o(X)$ to produce a classifier which is equivalent to the likelihood ratio test. A slight extension to the previous formulation can be enjoyed if we replace $\D(X)$ in \eqref{eq:old} by $\sigma(\D(X))$ where $\sigma(\cdot)$ is a scalar increasing function. In this more general setting belong [9] the regularized loss corresponding to $\big(1\pm\sigma(\D(X))\big)^2$ and the expectation loss corresponding to $|1\pm\sigma(\D(X))|$ (for binary classification the latter coincides with the Chebyshev loss [3], [7]) that accept optimum solutions which satisfy the basic requirement of being equivalent to LRT.

\subsection{Proposed optimization problems}
\vskip-0.2cm Let us now introduce our own class of problems. The difference between the methodology we intend to 
present and the existing techniques lies in the fact that in our case, provided certain very simple conditions are met, it is straightforward to show that the optimum is indeed a strategy equivalent to LRT. In place of \eqref{eq:old}, we propose the following criterion and the accompanying optimization
\begin{equation}
\mathcal{J}(\D)=\p_1\Exp_1[\phi\big(\D(X)\big)]-\p_2\Exp_2[\phi\big(\D(X)\big)],~~\max_{\D}\mathcal{J}(\D),
\label{eq:newnew}
\end{equation}
where $\phi(z)$ a scalar function. 

Using change of measure we can  rewrite our criterion as
\begin{multline}\textstyle
\mathcal{J}(\D)=\bar{\Exp}\left[\frac{\p_1\f_1(X)}{\p_1\f_1(X)+\p_2\f_2(X)}\phi\big(\D(X)\big)\right]-\bar{\Exp}\left[\frac{\p_2\f_2(X)}{\p_1\f_1(X)+\p_2\f_2(X)}\phi\big(\D(X)\big)\right]\\
=\bar{\Exp}[\big(\p_1(X)-\p_2(X)\big)\phi\big(\D(X)\big)],
\label{eq:nice11}
\end{multline}
where $\bar{\Exp}[\cdot]$ denotes expectation with respect to the mixture density $\bar{\f}(X)=\p_1\f_1(X)+\p_2\f_2(X)$ and $\p_k(X)=\frac{\p_k\f_k(X)}{\p_1\f_1(X)+\p_2\f_2(X)},k=1,2,$ is the \textit{posterior probability} that $X$ belongs to the class $\Hyp_k$. For the pair of functions $\phi(z),\D(X)$ we distinguish two categories.

\textbf{Category A.} 
Let $\phi(z)$ be a scalar function which satisfies
\begin{equation}
-1=\phi(-1)\leq\phi(z)\leq\phi(1)=1,~\text{for all real}~z.
\label{eq:new1}
\end{equation}
In other words $\phi(z)$ has a global minimum equal to $-1$ attained at $z=-1$ and a global maximum equal to 1 attained at $z=1$. The class of functions satisfying \eqref{eq:new1} is very rich. In fact if a scalar function $\varphi(z)$ has finite maximum and minimum attained at finite points $z_1,z_2$ then, with $\alpha,\beta,\gamma,\delta$ suitable constants, it can be transformed into a function $\phi(z)=\alpha\varphi(\gamma z+\delta)+\beta$ satisfying \eqref{eq:new1}. 

Any function $\phi(z)$ satisfying \eqref{eq:new1}, when used in \eqref{eq:newnew}, generates an optimization problem with solution equivalent to LRT. The validity of this claim is straightforward to demonstrate since from \eqref{eq:nice11}
\begin{equation*}
\mathcal{J}(\D)\leq\bar{\Exp}[|\p_1(X)-\p_2(X)|],
\end{equation*}
with the \textit{upper bound} being attainable by $\D_o(X)=\text{sign}\big(\p_1(X)-\p_2(X)\big)=\text{sign}\big(\p_1\f_1(X)-\p_2\f_2(X)\big)$, namely, a classifier which is equivalent to LRT.

\textbf{Category B.} 
Here $\phi(z)$ is assumed to be strictly increasing for $z\in[-1,1]$ with $\phi(1)=-\phi(-1)=1$. Actually, any strictly increasing function can be properly scaled to satisfy this constraint. Consider now the classifier function $\D(X)$ which we like to determine. In the previous category $\D(X)$ was not limited in any sense. In this case we impose the following boundedness condition
\begin{equation}
|\D(X)|\leq1.
\label{eq:constraint}
\end{equation}
If we attempt to solve \eqref{eq:newnew} then, again, it is straightforward to see that
\begin{equation*}
\mathcal{J}(\D)\leq\bar{\Exp}[|\p_1(X)-\p_2(X)|],
\end{equation*}
with the upper bound attained, as in the previous category, by $\D_o(X)=\text{sign}\big(\p_1(X)-\p_2(X)\big)=\text{sign}\big(\p_1\f_1(X)-\p_2\f_2(X)\big)$.

Even though we did not impose any additional conditions on $\phi(z)$ beyond the ones that define the two categories, it is understood that this function must be differentiable, except perhaps a finite number of points where it must have right and left derivatives. This is necessary in order to be able to derive, in the next section, gradient-type training algorithms. Regarding Category~A, it is preferable that $\phi(z)$ has no extra local extrema except the two global ones appearing at $z=\pm1$. This will help the training algorithm to avoid convergence to incorrect limits. Finally, we should mention that the log loss $\log\sigma(z)$ and the square log loss function $\big(\log\sigma(z)\big)^2$ [2] are special cases of Category~B. 

\section{Neural network classifiers}
\vskip-0.3cm Suppose now that we are interested in restricting the function $\D(X)$ to be \textit{the scalar output $\D(X,\theta)$ of a neural network} where $\theta$ summarizes the parameters of the network. The classifier and its error probability, following \eqref{eq:class}, become functions of $\theta$
\begin{equation}
\d(X,\theta)=\left\{\begin{array}{cl}
1,&\text{if}~\D(X,\theta)\geq0\\
2,&\text{if}~\D(X,\theta)<0,
\end{array}\right.
~~\text{and}~~\Pro_{\rm err}(\theta)=\p_1\Exp_1[\ind{\D(X,\theta)<0}]+\p_2\Exp_2[\ind{\D(X,\theta)\geq0}].
\label{eq:2}
\end{equation}
Similarly, for the cost proposed in \eqref{eq:newnew} we have
\begin{equation}
\mathcal{J}(\theta)=\p_1\Exp_1[\phi\big(\D(X,\theta)\big)]-\p_2\Exp_2[\phi\big(\D(X,\theta)\big)].
\label{eq:nice1}
\end{equation}
Maximization of $\mathcal{J}(\cdot)$ over the classifier is reduced to $\max_\theta \mathcal{J}(\theta)$, namely, maximization over the parameters of the network. This will produce an optimum neural network by identifying the best $\theta_o$. 
\vfill
\newpage

In the ideal case when $\D(X)$ is any arbitrary scalar function, the optimum solutions of all the alternative optimization problems \textit{are equivalent} since they match some version of LRT. Unfortunately, \textit{this significant property is lost} when we limit ourselves to neural networks. The optimum parameters $\theta_o$ and, therefore, the resulting network $\D(X,\theta_o)$ are \textit{optimization problem dependent}. 

The previous observation raises then the logical question: Since the possible ``optimum'' choices $\theta_o$ are not equal, which $\theta_o$ is the most appropriate for classification? Clearly, the answer is: \textit{the one that minimizes $\Pro_{\rm err}(\theta)$} defined in \eqref{eq:2}, because this produces the smallest misclassification probability. We stress again that the reason we do not perform the minimization of $\Pro_{\rm err}(\theta)$ and we resort to alternative problems is because of the presence of indicator functions that make it impossible to develop (stochastic) gradient type algorithms and iteratively compute the minimizer of $\Pro_{\rm err}(\theta)$.

\subsection{Data driven optimization problems}
\vskip-0.2cm The next step in our presentation consists in assuming that the probability densities $\f_1(X),\f_2(X)$ and the prior probabilities $\p_1,\p_2,$ are unknown. Instead, we are given two sets of data $\{X_1^1,\ldots,X_{N_1}^1\}$ and $\{X_1^2,\ldots,X_{N_2}^2\}$ which are realizations that follow the two unknown densities. Furthermore, for the sizes of the two data sets $N_1,N_2,$ we assume that they are consistent with the two unknown prior probabilities $\p_1,\p_2$ in the sense that $\p_k\approx\frac{N_k}{N_1+N_2},~k=1,2$. 

Computing the two expectations in \eqref{eq:nice1} is no longer an option, hence, it makes sense to approximate stochastic means by sample averages with the help of the available data. More precisely
\begin{multline}\allowdisplaybreaks
\textstyle
\mathcal{J}(\theta)\approx\hat{\mathcal{J}}(\theta)=\frac{N_1}{N_1+N_2}\left\{\frac{1}{N_1}\sum_{i=1}^{N_1}\phi\big(\D(X_i^1,\theta)\big)\right\}-\frac{N_2}{N_1+N_2}\left\{\frac{1}{N_2}\sum_{i=1}^{N_2}\phi\big(\D(X_i^2,\theta)\big)\right\}\allowdisplaybreaks\\
\textstyle
=\frac{1}{N_1+N_2}\left\{\sum_{i=1}^{N_1}\phi\big(\D(X_i^1,\theta)\big)-\sum_{i=1}^{N_2}\phi\big(\D(X_i^2,\theta)\big)\right\}.\allowdisplaybreaks
\label{eq:nice0}
\end{multline}
It is then clear that
\begin{equation}
\max_\theta \hat{\mathcal{J}}(\theta)
\label{eq:nice3}
\end{equation}
replaces the maximization of $\mathcal{J}(\theta)$. We note that for the computation of $\hat{\mathcal{J}}(\theta)$ in \eqref{eq:nice0} we need the training data, the function $\phi(z)$ and the geometry of the neural network. No knowledge or modeling of densities or priors is necessary. We should also mention that if we perform a similar approximation for $\Pro_{\rm err}(\theta)$ we obtain
\begin{multline}
\textstyle
\Pro_{\rm err}(\theta)\approx\hat{\Pro}_{\rm err}(\theta)=\frac{1}{N_1+N_2}\left\{\sum_{i=1}^{N_1}\ind{\D(X_i^1,\theta)<0}
+\sum_{i=1}^{N_2}\ind{\D(X_i^2,\theta)\geq0}\right\}\\
\textstyle
=\frac{N_1}{N_1+N_2}-\frac{1}{N_1+N_2}\left\{\sum_{i=1}^{N_1}\ind{\D(X_i^1,\theta)\geq0}
-\sum_{i=1}^{N_2}\ind{\D(X_i^2,\theta)\geq0}\right\},
\label{eq:nice10}
\end{multline}
which constitutes the preferable data driven criterion to optimize.

\textbf{Summary.} \textit{For the design of the neural network classifier, we propose the solution of the optimization problem depicted in \eqref{eq:nice3}. For the function $\phi(z)$ and the output $\D(X,\theta)$ of the neural network we offer two possibilities. \textit{\textbf{Category~A:}} $\phi(z)$ must have a global minimum equal to $-1$ at $z=-1$ and a global maximum equal to 1 at $z=1$. No condition is imposed on the output $\D(X,\theta)$ of the neural network. \textit{\textbf{Category~B:}} $\phi(z)$ must be increasing in $[-1,1]$ with $\phi(1)=1$ and $\phi(-1)=-1$. The output $\D(X,\theta)$ of the neural network must be limited within the interval $[-1,1]$.}

\begin{figure}[!h]
\centering
\vskip-0.cm
\hbox to \hsize{\hfill\includegraphics[width=0.35\hsize]{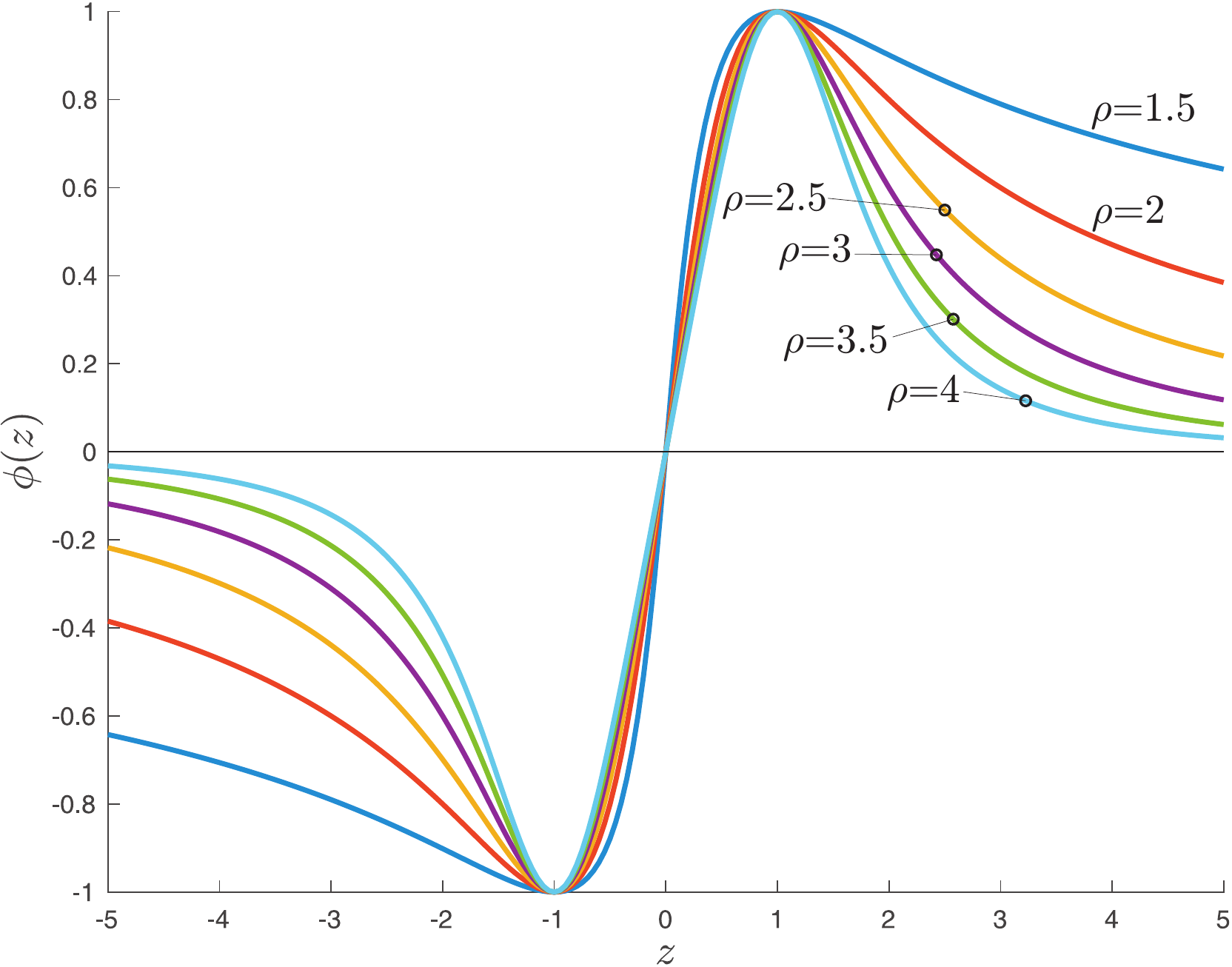}\hfill\includegraphics[width=0.35\hsize]{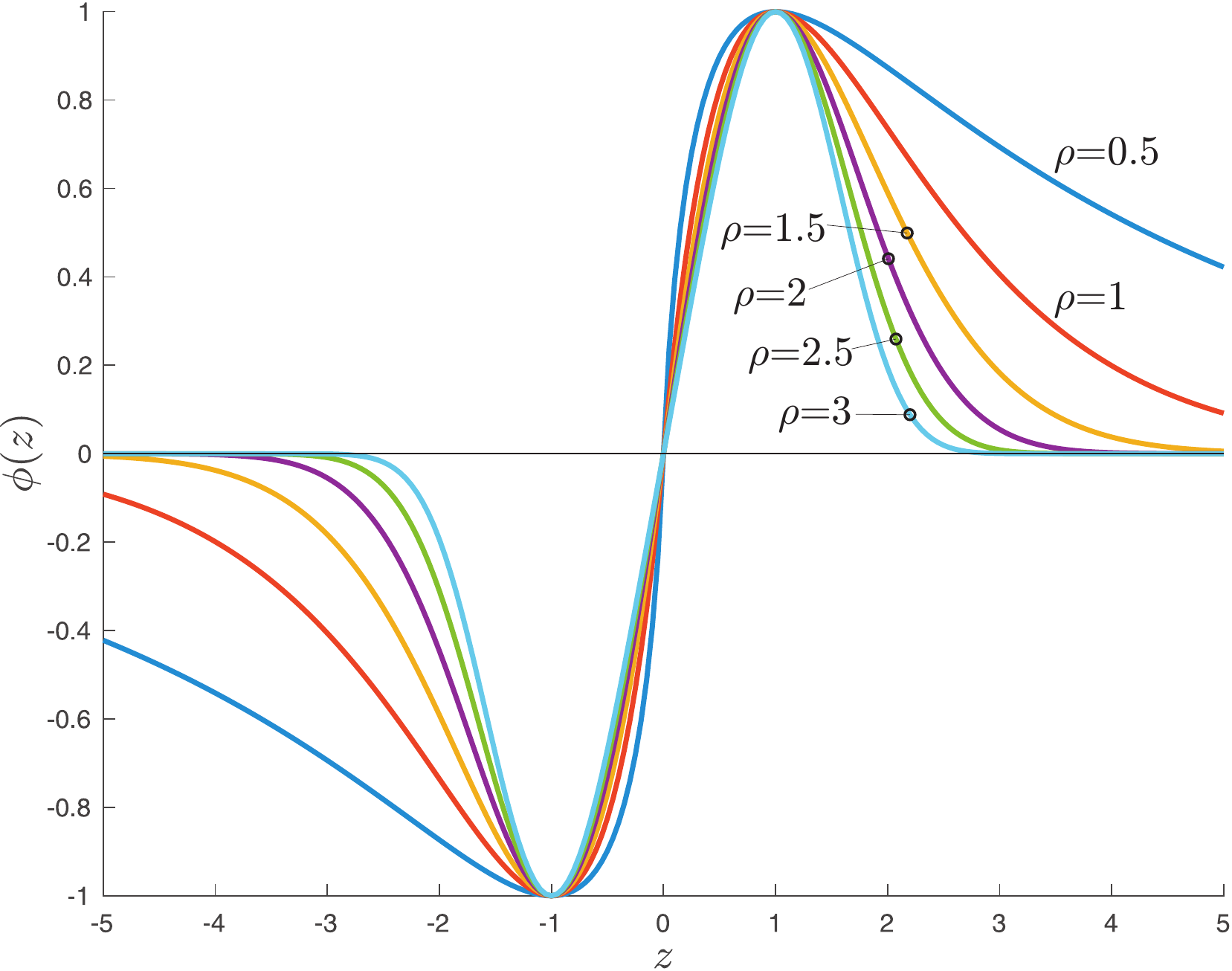}\hfill}
\vskip-0.15cm
\hbox to \hsize{\hfill\hbox to 0.35\hsize{\hfil\small(a)\hfil}\hfill\hbox to 0.35\hsize{\hfil\small(b)\hfil}\hfill}
\vskip-0.2cm
\caption{Examples for $\phi(z)$: (a)~$\phi(z)=\frac{\rho z}{\rho-1+|z|^\rho}$;
(b)~$\phi(z)=ze^{\frac{1}{\rho}(1-|z|^\rho)}$.}
\label{fig:1}
\end{figure}
\newpage
For Category~A, potential forms for $\phi(z)$ are 
\begin{equation*}
\phi(z)=\frac{\rho z}{\rho-1+|z|^\rho},\rho>1;~~~\text{or}~~\phi(z)=ze^{\frac{1}{\rho}(1-|z|^\rho)},\rho>0.
\end{equation*}
The two possibilities are depicted in Figure\,\ref{fig:1} for different values of the parameter $\rho$. As we can see, in the first case the tails of the function decrease to 0 as $\frac{1}{z}$ raised to some fixed power. This suggests that $\phi(z)$ has ``fat'' tails, which in turn implies that outputs of the network that are far from the two target values $\pm1$ can still contribute to the overall criterion in \eqref{eq:nice0}. If on the other hand we use the second function where tails decrease to zero rapidly then, network outputs far from the target values tend to have small or even negligible contribution to the overall cost. This form of ``data screening'' could, potentially, be advantageous when robustness against possible outliers is necessary.

\subsection{Training algorithms}
\vskip-0.2cm Let us now continue with a detailed presentation of possible training algorithms for the optimization problem introduced in \eqref{eq:nice3}. We limit ourselves to a full, 2-layer network, because in most of our simulations we observed satisfactory performance from this simple geometry. Of course one can very easily extend our derivations to cover networks with more layers and/or with special structure as convolutional neural networks. 

For Category~A, we recall that there is no nonlinear transformation in the last layer of the neural network. On the output, we do however apply the nonlinearity $\phi(z)$ which we select for our criterion in \eqref{eq:nice0}. For Category~B because of the constraint in \eqref{eq:constraint} we need to apply some nonlinearity $g(z)$ to contain the network output in the interval $[-1,1]$. On top of this nonlinearity, according to our approach, we must apply the strictly increasing function $\phi(z)$ that we select for our optimization problem. 

Both categories can be put under the same form, allowing for the presentation of a common training algorithm. For this reasons, let $X$ denote the length $k$ vector that we like to classify. We then apply the following transformations
\begin{equation}
U=\pzA X+\pa,~Z=d(U),~z=\pB^{\intercal}Z+\pb,~y=\omega(z),
\label{eq:nn}
\end{equation}
where $A$ is an $n\times k$ dimensional matrix, $\{\pa,U,Z,\pB\}$ are vectors of length $n$, $d(x)$ is one of the popular scalar nonlinearities employed in neural networks and applied to each element of the vector $U$ and $\pb$ is a scalar offset. For the output nonlinearity we have for Category~A that $\omega(z)=\phi(z)$ while for Category~B the nonlinearity takes the form $\omega(z)=(\phi\circ g)(z)$ (composition of the two functions). Clearly the scalar output $y$ takes over the role of $\D(X,\theta)$ and $\theta$ summarizes the quantities $\{\pzA,\pa,\pB,\pb\}$ which need to be identified.

To develop a training algorithm that solves \eqref{eq:nice3} we must find the gradient of the cost function with respect to all network parameters. This translates into computing the gradient of $\omega(z)$ with respect to $\pzA,\pa,\pB,\pb$. We have the following formulas that can be easily verified
\begin{gather}
\pH=\nabla_{\!\!\pzA}\omega(z)=\omega'(z)\big(\pB \odot d'(U)\big)X^{\intercal},~~\ph=\nabla_{\!\!\pa}\omega(z)=\omega'(z)\big(\pB \odot d'(U)\big)\nonumber\\
\pG=\nabla_{\!\!\pB}\omega(z)=\omega'(z)Z,~~\pg=\nabla_{\!\pb}\omega(z)=\omega'(z),\nonumber
\end{gather}
where $\odot$ denotes element-by-element multiplication of the corresponding vectors, ``$^\intercal$'' denotes transpose and ``\,$'$\,'' derivative. For the computation of the solution of the optimization problem in \eqref{eq:nice3}, we distinguish a batch and a stochastic gradient version.

\textbf{Batch version.} We form a gradient ascent scheme by considering directly the criterion in \eqref{eq:nice0}. Table\,\ref{tab:1} summarizes the algorithm. In the corresponding formulas ``$(\cdot)^{(a)}$'' denotes the element-wise raise to the power $a$ and ``$\odiv$'' the element-by-element division of the corresponding vectors or matrices. Parameter $\mu$ is the learning rate and must be sufficiently small so that the algorithm does not diverge. Finally, $\lambda$ is an exponential forgetting factor and is used to estimate average powers of the gradient elements. Following the scheme in [13] we normalize each gradient element with the square root of its estimated power before using it in the update of the corresponding parameter. The batch version tends to become computationally demanding when $N_1,N_2$ are very large since it requires a number of computations per iteration which is proportional to $N_1+N_2$.
\begin{table}\small
\caption{Batch version}
\label{tab:1}
\vskip-0.2cm
\centering
\begin{tabular}{l}
\toprule
Initialize~$\pzA_0,\pB_0$ using the method in [8] and set $\pa_0=0,\pb_0=0,$\\
$\pM_0=0,\pzm_0=0,\pN_0=0,\pn_0=0$\\
\addlinespace[-1pt]
\midrule
Available from iteration $t-1$:~$\pzA_{t-1},\pa_{t-1},\pB_{t-1},\pb_{t-1},\pM_{t-1},\pzm_{t-1},\pN_{t-1},\pn_{t-1}$\\
\addlinespace[-1pt]
\midrule
At iteration $t$ and for $i=1,\ldots,N_k,~k=1,2$:\\
\addlinespace[1pt]
Compute layer outputs for all available data vectors:\\
~~~~$U_{t,i}^k=\pzA_{t-1}X^k_i+\pa_{t-1}$,~~$Z_{t,i}^k=d(U_{t,i}^k)$,~~$z_{t,i}^k=\pB^{\intercal}_{t-1}Z_{t,i}^k+\pb_{t-1}$\\
\addlinespace[1pt]
Compute gradients:\\
~~~~$\pH_t^k=\sum_{i=1}^{N_k}\omega'(z_{t,i}^k)\big(\pB_{t-1} \odot d'(U_{t,i}^k)\big)(X_i^k)^{\intercal},~~\ph_t^k=\sum_{i=1}^{N_k}\omega'(z_{t,i}^k)\big(\pB_{t-1} \odot d'(U_{t,i}^k)\big)$\\
\addlinespace[1pt]
~~~~$\pG_t^k=\sum_{i=1}^{N_k}\omega'(z_{t,i}^k)Z_{t,i}^k,~~\pg_t^k=\sum_{i=1}^{N_k}\omega'(z_{t,i}^k)$\\
\addlinespace[1pt]
Update power estimates:\\
~~~~$\pM_t=\lambda\pM_{t-1}+(1-\lambda)(\pH_t^1-\pH_t^2)^{(2)},~~\pzm_t=\lambda\pzm_{t-1}+(1-\lambda)(\ph_t^1-\ph_t^2)^{(2)}$\\
\addlinespace[1pt]
~~~~$\pN_t=\lambda\pN_{t-1}+(1-\lambda)(\pG_t^1-\pG_t^2)^{(2)},~~\pn_t=\lambda\pn_{t-1}+(1-\lambda)(\pg_t^1-\pg_t^2)^{(2)}$\\
\addlinespace[1pt]
Update parameter estimates:\\
~~~~$\pzA_t=\pzA_{t-1}+\mu\{\pH_t^1-\pH_t^2\}\odiv(\pM_t)^{(\frac{1}{2})}$,~~$\pa_t=\pa_{t-1}+\mu\{\ph_t^1-\ph_t^2\}\odiv(\pzm_t)^{(\frac{1}{2})}$\\
\addlinespace[1pt]
~~~~$\pB_t=\pB_{t-1}+\mu\{\pG_t^1-\pG_t^2\}\odiv(\pN_t)^{(\frac{1}{2})}$,~~$\pb_t=\pb_{t-1}+\mu\{\pg_t^1-\pg_t^2\}\odiv(\pn_t)^{(\frac{1}{2})}$\\
\midrule
Repeat until some stopping rule is satisfied\\
\bottomrule
\end{tabular}
\end{table}

\begin{table}[!h]\small
\caption{Stochastic Gradient version}
\label{tab:2}
\vskip-0.2cm
\centering
\begin{tabular}{l}
\toprule
Initialize~$\pzA_0,\pB_0$ using the method in [8] and set $\pa_0=0,\pb_0=0,$\\
$\pM_0=0,\pzm_0=0,\pN_0=0,\pn_0=0$\\
\addlinespace[-1pt]
\midrule
Available from iteration $t-1$:~$\pzA_{t-1},\pa_{t-1},\pB_{t-1},\pb_{t-1},\pM_{t-1},\pzm_{t-1},\pN_{t-1},\pn_{t-1}$\\
\addlinespace[-1pt]
\midrule
At iteration $t$ select the next data vector $X_t$ from the merged set (recycle when data~~~~~~~~~~~~~\\
are exhausted). Perform the following computations:\\
\addlinespace[1pt]
Compute layer outputs:\\
~~~~$U_t=\pzA_{t-1}X_t+\pa_{t-1}$,~~$Z_t=d(U_t)$,~~$z_t=\pB^{\intercal}_{t-1}Z_t+\pb_{t-1}$\\
\addlinespace[1pt]
Compute gradients:\\
~~~~$\pH_t=\omega'(z_t)\big(\pB_{t-1}\odot d'(U_t)\big)X_t^{\intercal},~~\ph_t=\omega'(z_t)\big(\pB_{t-1} \odot d'(U_t)\big)$\\
\addlinespace[1pt]
~~~~$\pG_t=\omega'(z_t)Z_t,~~\pg_t=\phi'(z_t)$\\
\addlinespace[1pt]
Update power estimates:\\
~~~~$\pM_t=\lambda\pM_{t-1}+(1-\lambda)(\pH_t)^{(2)},~~\pzm_t=\lambda\pzm_{t-1}+(1-\lambda)(\ph_t)^{(2)}$\\
\addlinespace[1pt]
~~~~$\pN_t=\lambda\pN_{t-1}+(1-\lambda)(\pG_t)^{(2)},~~\pn_t=\lambda\pn_{t-1}+(1-\lambda)(\pg_t)^{2}$\\
\addlinespace[1pt]
Update parameter estimates:\\
~~~~$\epsilon_t=1$ if label of $X_t$ is 1 and $\epsilon_t=-1$ if label of $X_t$ is 2\\
\addlinespace[1pt]
~~~~$\pzA_t=\pzA_{t-1}+\mu\epsilon_t\pH_t\odiv(\pM_t)^{(\frac{1}{2})}$,~~$\pa_t=\pa_{t-1}+\mu\epsilon_t\ph_t\odiv(\pzm_t)^{(\frac{1}{2})}$\\
\addlinespace[1pt]
~~~~$\pB_t=\pB_{t-1}+\mu\epsilon_t\pG_t\odiv(\pN_t)^{(\frac{1}{2})}$,~~$\pb_t=\pb_{t-1}+\mu\epsilon_t\pg_t\odiv(\pn_t)^{(\frac{1}{2})}$\\
\midrule
Repeat until some stopping rule is satisfied\\
\bottomrule
\end{tabular}
\vskip-0.5cm
\end{table}
\textbf{Stochastic gradient version.} For this version we need the two data sets to be merged into a single set and \textit{the data to be randomly permuted}. In the merged set the data must of course retain their original labeling. Table\,\ref{tab:2} summarizes the algorithm. We would like to emphasize that the random permutation of the data is absolutely necessary because, otherwise, if the data are grouped according to their labels, the algorithm will exhibit a periodically biased convergence behavior as the data are being reused. In fact, it would be advisable to perform a new random permutation every time we recycle the data.

\textbf{Remark.} \textit{Regarding the nonlinearity $d(x)$ applied to the output of the first layer, simulations agree that the best choice is to use the ReLU since the resulting algorithm exhibits a far more stable convergence behavior as compared to other alternatives.}

\section{Simulations}
\vskip-0.4cm To test our methodology, from Category~A we select $\phi(z)=\frac{2z}{1+z^2}$, while from Category~B we use $\phi(z)=z$ and for limiting the output of the neural network in $[-1,1]$ we adopt $g(z)=\tanh(z)$. We would like to add that we also performed simulations with $\phi(z)=ze^{1-|z|}$ from Category~A. However, the results were almost identical to $\phi(z)=\frac{2z}{1+z^2}$, for this reason we are not including them in our presentation. We compare our algorithms against the method based on the Hinge loss [12]. In [9] this technique was evaluated and found to enjoy many positive characteristics compared to other possibilities. 

Before presenting our simulation results we would like to mention a very interesting interpretation for our Category~B selection. As we claimed before, the most desirable criterion to minimize is $\hat{\Pro}_{\rm err}(\theta)$. It is because of the indicator function that we seek different formulations. If in \eqref{eq:nice10} we \textit{approximate} the indicator $\ind{z\geq0}$ with the sigmoid $\frac{1}{1+e^{-2z}}=0.5\{\tanh(z)+1\}$ then, minimizing the resulting criterion is equivalent to maximizing the part inside the brackets, which is exactly our optimization problem in \eqref{eq:nice3} for our Category~B selection. This connection to the desired optimization problem will prove to be beneficial for our method as we will see next.

In the first set of experiments we considered scalar random variables ($k=1$) and attempted to classify between a standard normal $\mathcal{N}(0,1)$ under $\Hyp_1$ and a mixture of Gaussians $0.6\mathcal{N}(1,1)+0.4\mathcal{N}(-3,1)$ under $\Hyp_2$. We used a two-layer network with the first layer output having length $n=100$ and we applied the ReLU nonlinearity $d(z)=(z)^+$. In all three training algorithms we used the same learning rate $\mu=10^{-4}$ and forgetting factor $\lambda=0.99$. The networks were initialized with exactly the same values following the initialization scheme in [8] and the gradients were normalized following the scheme in [13]. 

Regarding the training data, we used $N_1=N_2=5000$ and applied the stochastic gradient algorithm. To avoid the need for random permutations, in each iteration we simply used one data point from each class, instead of randomly switching between classes. To evaluate the quality of the corresponding network, \textit{at each iteration} after the parameters were updated, we applied the resulting network to $10^5$ testing data from $\Hyp_1$ and an equal number from $\Hyp_2$ in order to estimate the corresponding error probabilities the classifiers could deliver. There are two error types, one for each data class and there is also their average which, as we know, is optimized by LRT.
\begin{figure}[!h]
\centering
\hbox to \hsize{\hfill\includegraphics[width=0.326\hsize]{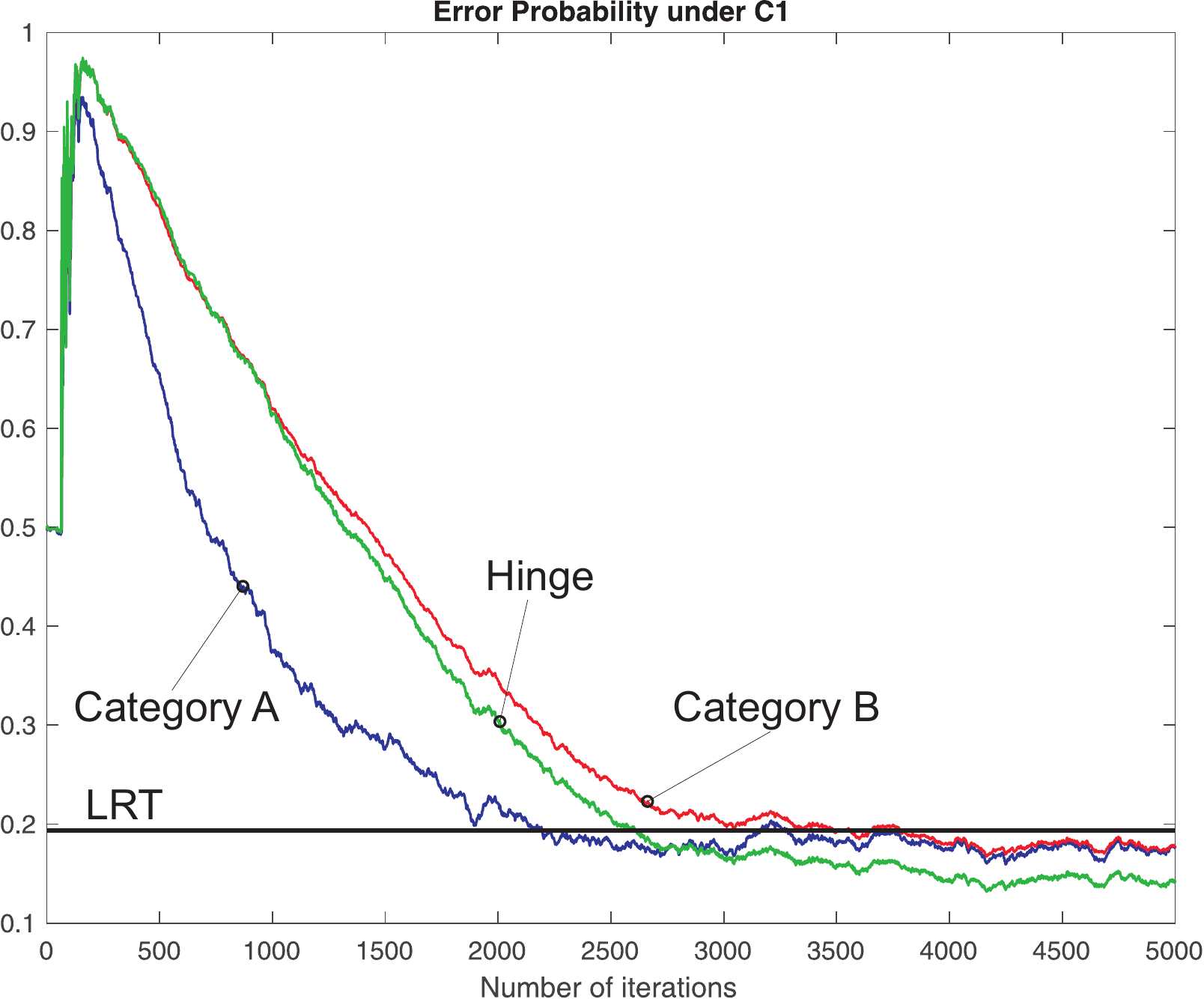}\hfill\includegraphics[width=0.33\hsize]{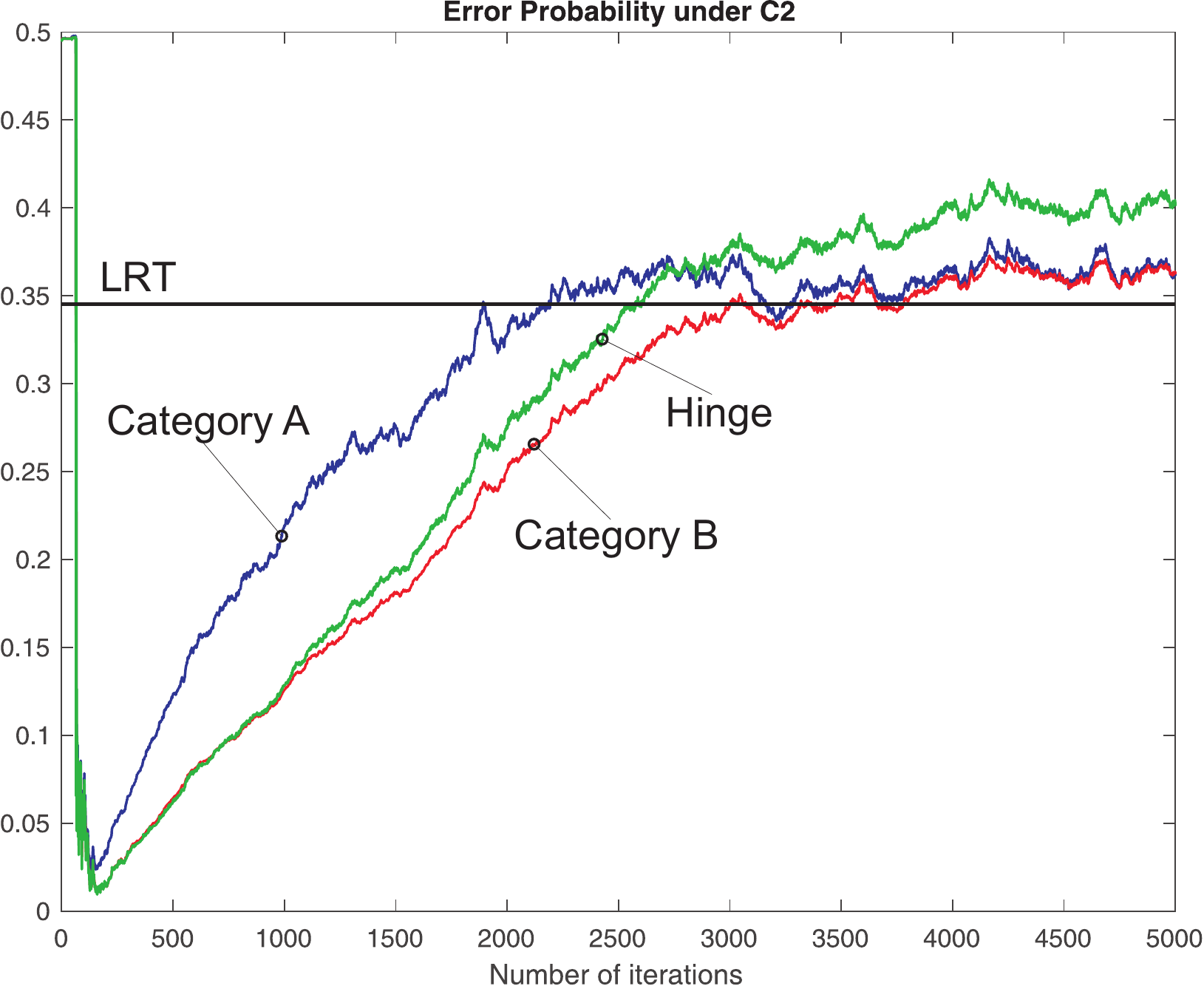}\hfill\includegraphics[width=0.33\hsize]{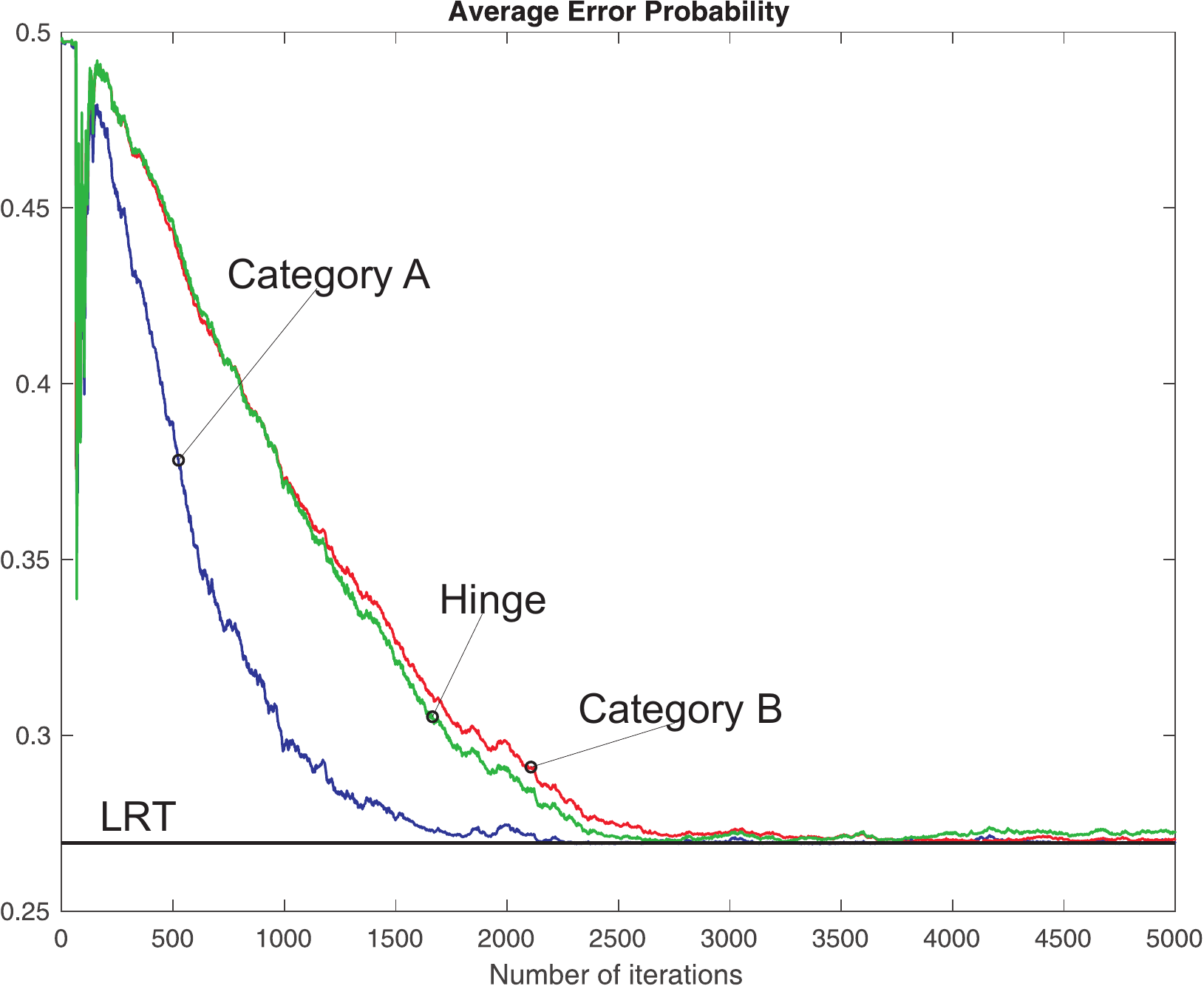}\hfill}
\hbox to \hsize{\hfill\hbox to 0.326\hsize{\hfil{\small(a)}\hfil}\hfill\hbox to 0.33\hsize{\hfil\small(b)\hfil}\hfill\hbox to 0.33\hsize{\hfil\small(c)\hfil}\hfill}
\vskip-0.1cm
\caption{Evolution of classification error probabilities with number of iterations.}
\label{fig:2}
\vskip-0.3cm
\end{figure}

In Figure\,\ref{fig:2} we plot the evolution of the error probabilities as a function of the number of iterations of the training algorithms. We also include the errors delivered by LRT (for which we used the true densities). In Figure\,\ref{fig:2}(a) we present the errors when the data come from $\Hyp_1$, in Figure\,\ref{fig:2}(b) when the data come from $\Hyp_2$, and in Figure\,\ref{fig:2}(c) the average of the two errors. 

We have the following interesting observations: Our training algorithms can design neural networks that have classification errors that approach the errors of LRT more efficiently than the algorithm which is based on the Hinge loss. Remarkably, despite the significant differences, when the errors are averaged, all three classifiers yield similar performances, closely matching the optimum LRT performance. To obtain a more precise idea  about the corresponding errors, in Table\,\ref{tab:4} we present the error probabilities delivered by the networks as designed in the final (5000th) iteration.
\begin{table}[h!]
\small
\caption{Error Probabilities}
\label{tab:4}
\vskip-0.2cm
\centering
\begin{tabular}{lccc}
\toprule
Method&Error under $\Hyp_1$&Error under $\Hyp_2$&Average Error\\
\addlinespace[-2pt]
\midrule
LRT&0.194&0.345&0.269\\
\addlinespace[-2pt]
\midrule
Category~A&0.178&0.361&0.269\\
\addlinespace[-2pt]
\midrule
Category~B&0.178&0.362&0.270\\
\addlinespace[-2pt]
\midrule
Hinge&0.143&0.401&0.272\\
\addlinespace[-2pt]
\bottomrule
\end{tabular}
\end{table}
We can see that both our methods approximate the LRT errors better than the Hinge loss based scheme. The average error on the other hand for all three methods is extremely close the the optimum LRT performance. This specific behavior \textit{is typical} in all simulation we performed, where we experimented with numerous data dimensions, means, variances and mixture probabilities for the corresponding mixture densities.


The next simulation involves real datasets. Here, we limit ourselves to the Category~A algorithm (since it has shown slightly better convergence speed than its Category~B counterpart) and we compare it against the Hinge loss based scheme. The goal is to distinguish between the two hand written numerals ``4'' and ``9'' from the MNIST database. 

\begin{figure}[!h]
\centering
\hbox to \hsize{\hfill\includegraphics[width=0.33\hsize]{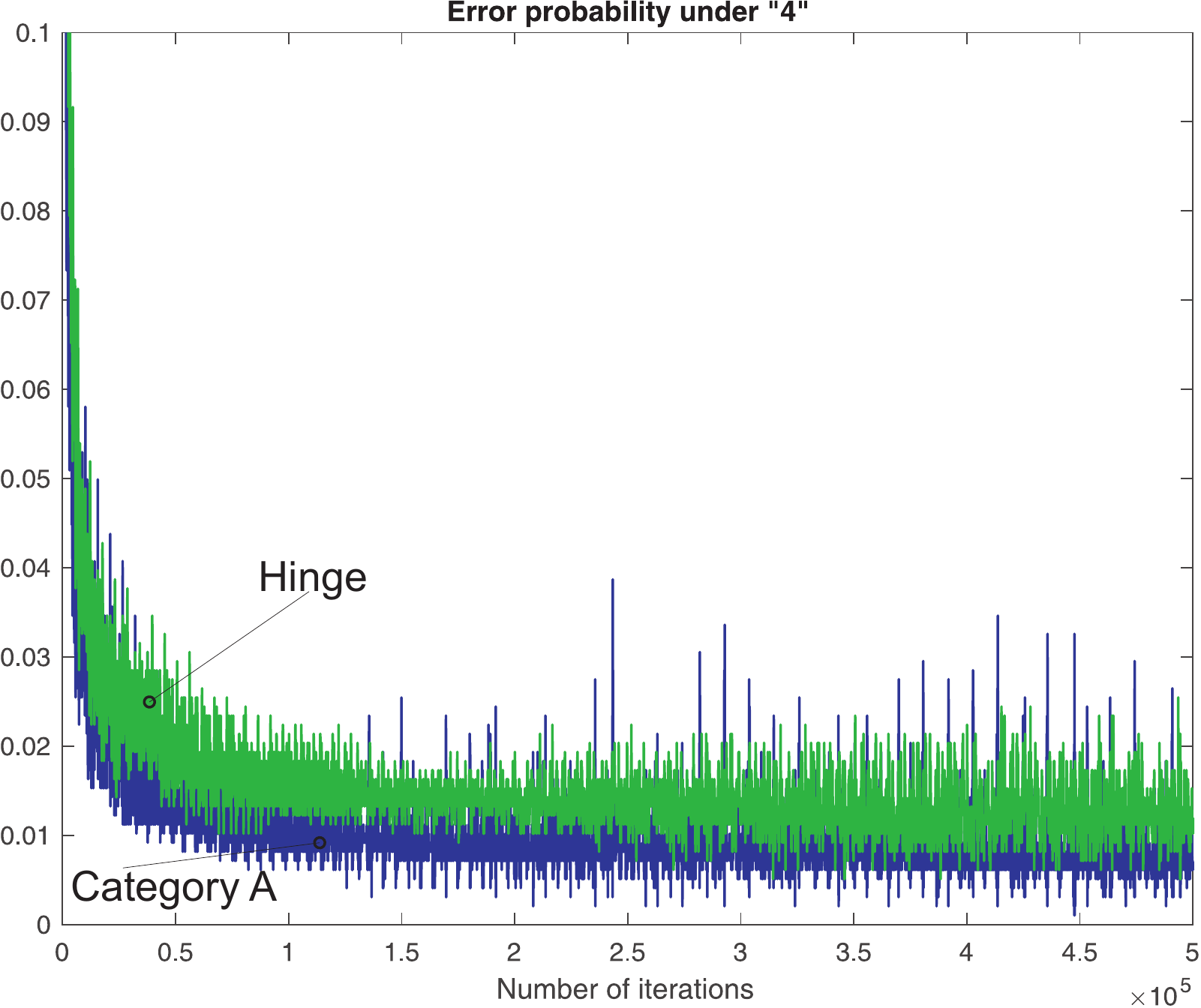}\hfill\includegraphics[width=0.33\hsize]{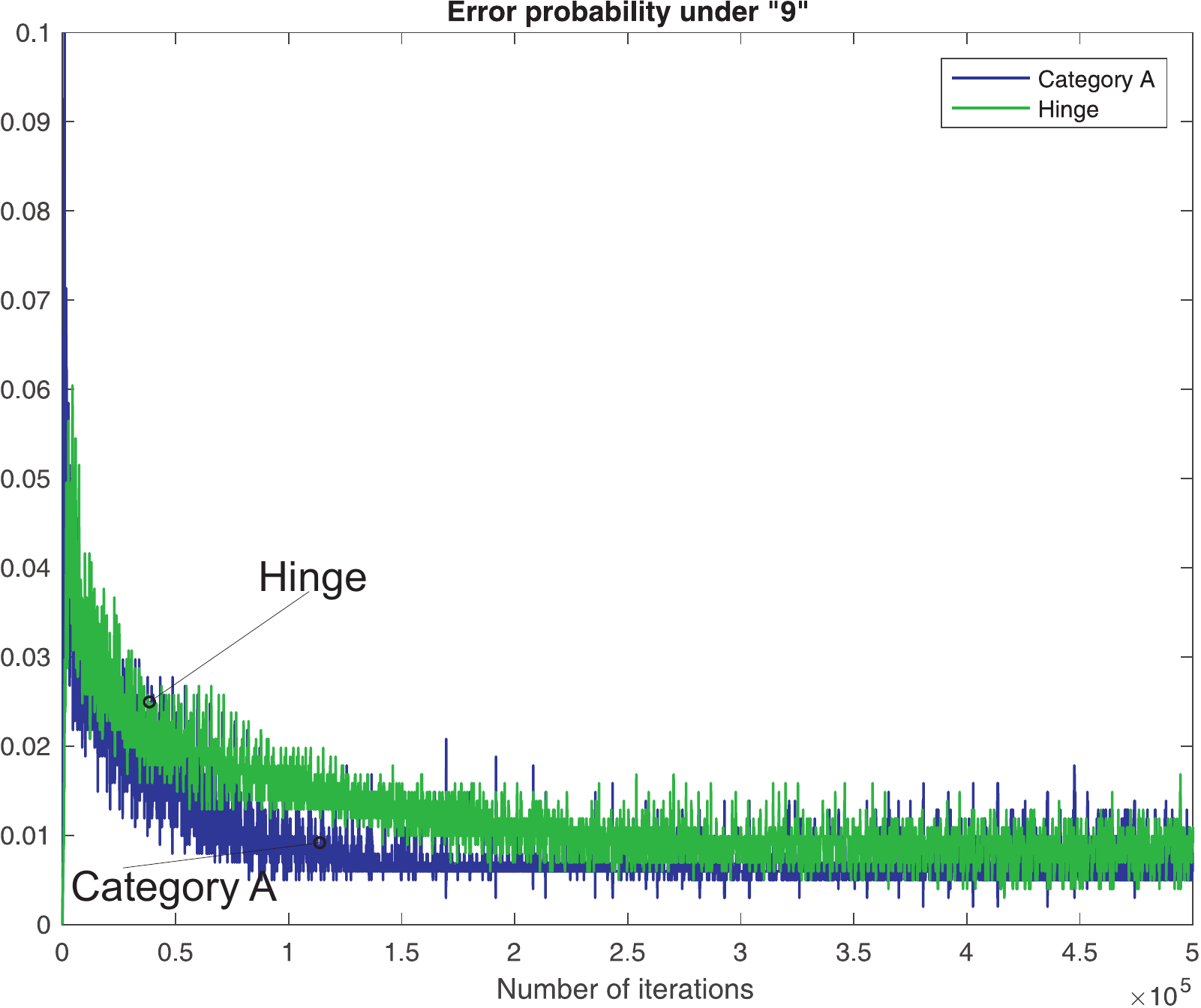}\hfill\includegraphics[width=0.33\hsize]{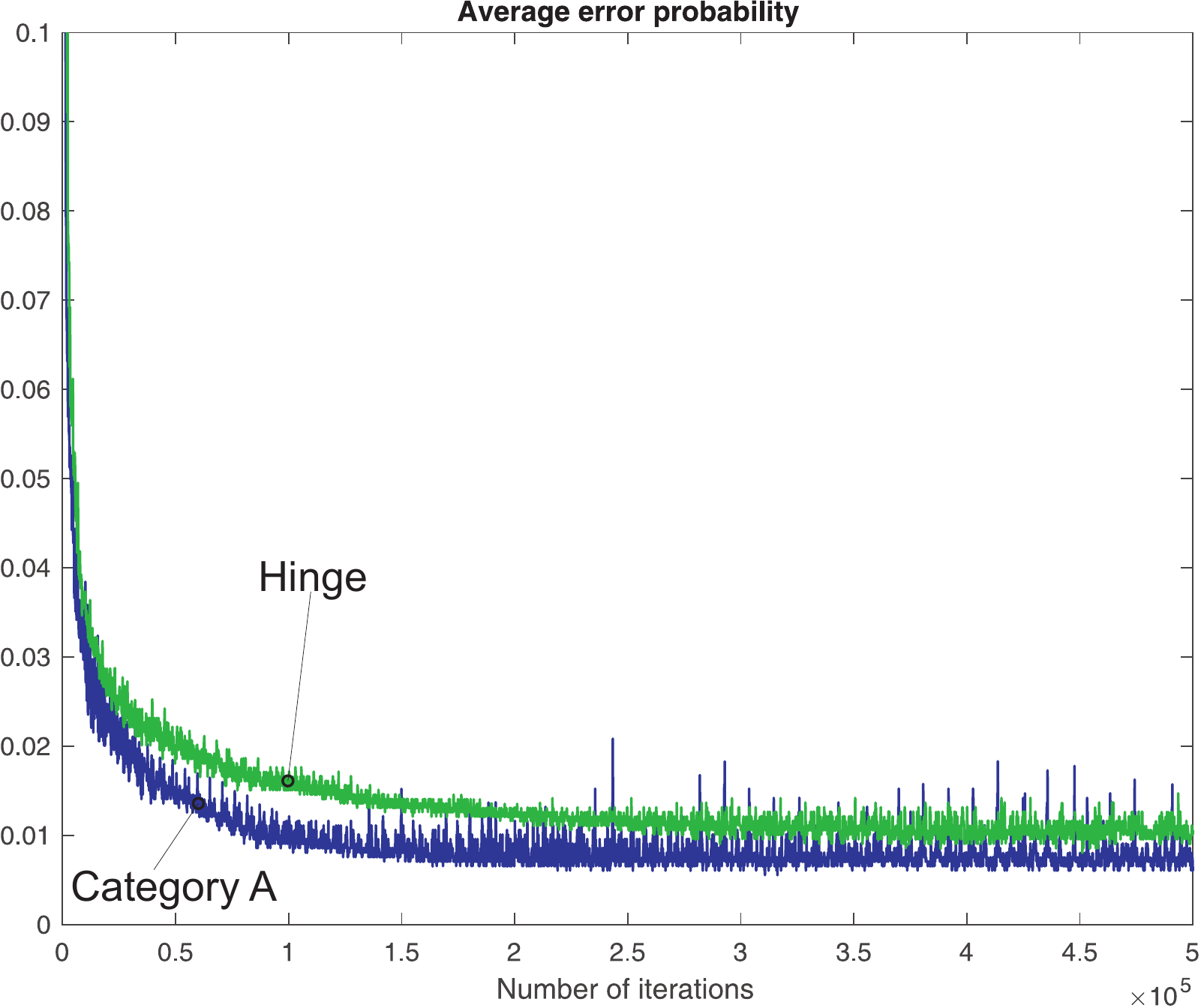}\hfill}
\hbox to \hsize{\hfill\hbox to 0.33\hsize{\hfil{\small(a)}\hfil}\hfill\hbox to 0.33\hsize{\hfil\small(b)\hfil}\hfill\hbox to 0.33\hsize{\hfil\small(c)\hfil}\hfill}
\vskip-0.1cm
\caption{Evolution of classification error probabilities with number of iterations for the handwritten MNIST numerals ``4'' and ``9''.}
\label{fig:3}
\end{figure}

The image size is $28\times28$ and is reshaped into a vector of length $k=784$. We use a full two layer network with the first layer output being of length $n=300$. The learning rates are selected equal to $\mu=10^{-4}$ and the forgetting factors equal to $0.99$. Our training set is comprised of $N_1=5500$ handwritten ``4'' and $N_2=5500$ handwritten ``9''. We also have an additional 982 handwritten ``4'' and 1009 handwritten ``9'' that we use for testing. Every time the training data are exhausted we recycle them. Following the same strategy as in the previous experiment, at each iteration we test the quality of the computed classifiers by applying them to the testing data. Figure\,\ref{fig:3}, similarly to Figure\,\ref{fig:2}, depicts the evolution of the two errors and their average as a function of the number of iterations.
As we can see our scheme exhibits a faster convergence rate and it attains lower levels of average error probability. If we now focus on the neural networks obtained during the last ($5\times10^5$th) iteration then our design method applied to the testing data makes 4 errors out of 982 when applied to the set of ``4'' and 12 errors out of 1009 when applied to the set of ``9''. The average number of errors is 8. The same figures for the Hinge based scheme are 17 and 5 with average equal to 11.

\begin{figure}[!h]
\centering
\hbox to \hsize{\hfill\includegraphics[width=0.25\hsize,height=0.8cm]{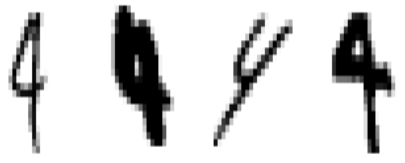}\hfill\includegraphics[width=0.25\hsize,height=0.8cm]{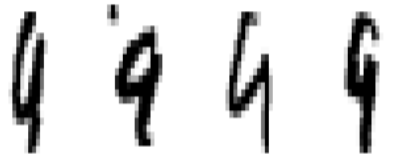}\hfill}
\vskip-0.1cm
\hbox to \hsize{\hfill\hbox to 0.25\hsize{\hfil{\small(a)}\hfil}\hfill\hbox to 0.25\hsize{\hfil\small(b)\hfil}\hfill}
\vskip0.2cm
\hbox to \hsize{\hfill\includegraphics[width=0.25\hsize,height=0.8cm]{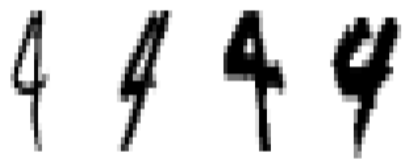}\hfill\includegraphics[width=0.25\hsize,height=0.8cm]{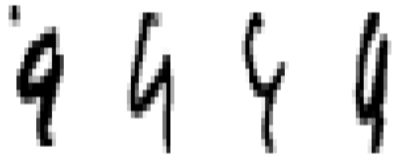}\hfill}
\vskip-0.1cm
\hbox to \hsize{\hfill\hbox to 0.25\hsize{\hfil{\small(c)}\hfil}\hfill\hbox to 0.25\hsize{\hfil\small(d)\hfil}\hfill}
\vskip-0.1cm
\caption{Examples of classification errors. Proposed, misclassifying (a) ``4'' as ``9'' and (b) ``9'' as ``4''; Hinge based, misclassifying (c) ``4'' as ``9''and (d) ``9'' as ``4''.}
\label{fig:4}
\vskip-0.2cm
\end{figure}

It is interesting to visualize a few examples where each method fails. In Figure\,\ref{fig:4} we depict four errors for each method and each data class. Figure\,\ref{fig:4}(a) contains cases where our scheme misclassified ``4'' as ``9'' and in (b) the opposite, namely, ``9'' as ``4''. In Figure\,\ref{fig:4}(c) and (d) we have the corresponding failures for the Hinge based method. We could agree that in most of these cases the correct decision would have been challenging even for a human decision maker.

A number of interesting experiments and comparisons follow based on the CIFAR-10 database where we tested several combinations of pairs of classes. This particular set of experiments involves far less (statistically) structured data than the preceding examples therefore the error probabilities we obtain are significantly more modest. We focus in comparing our Category B algorithmic version with the classical Hinge loss based algorithm. Each class contains 5000 training data and 1000 testing data. In Figures 5, 6, 7, 8 and 9 we present, as before, the evolution of the classification error probabilities with the number of iterations for the pairs ``Cats \& Dogs'', ``Airplanes \& Automobiles'', ``Deers \& Birds'', ``Frogs \& Horses'', ``Ships \& Trucks'' respectively.

\begin{figure}[!p]
\centering
\hbox to \hsize{\hfill\includegraphics[width=0.33\hsize]{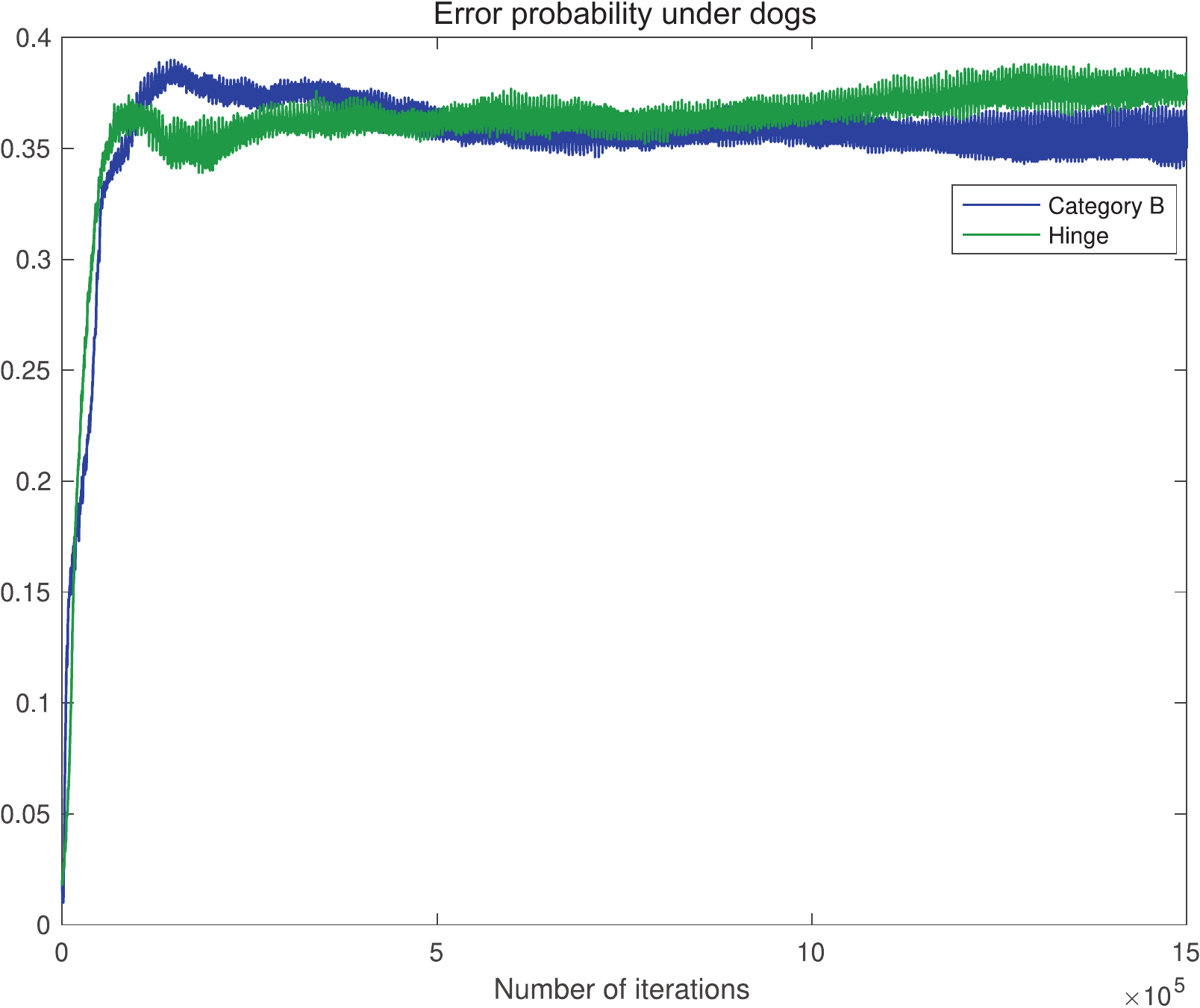}\hfill\includegraphics[width=0.325\hsize]{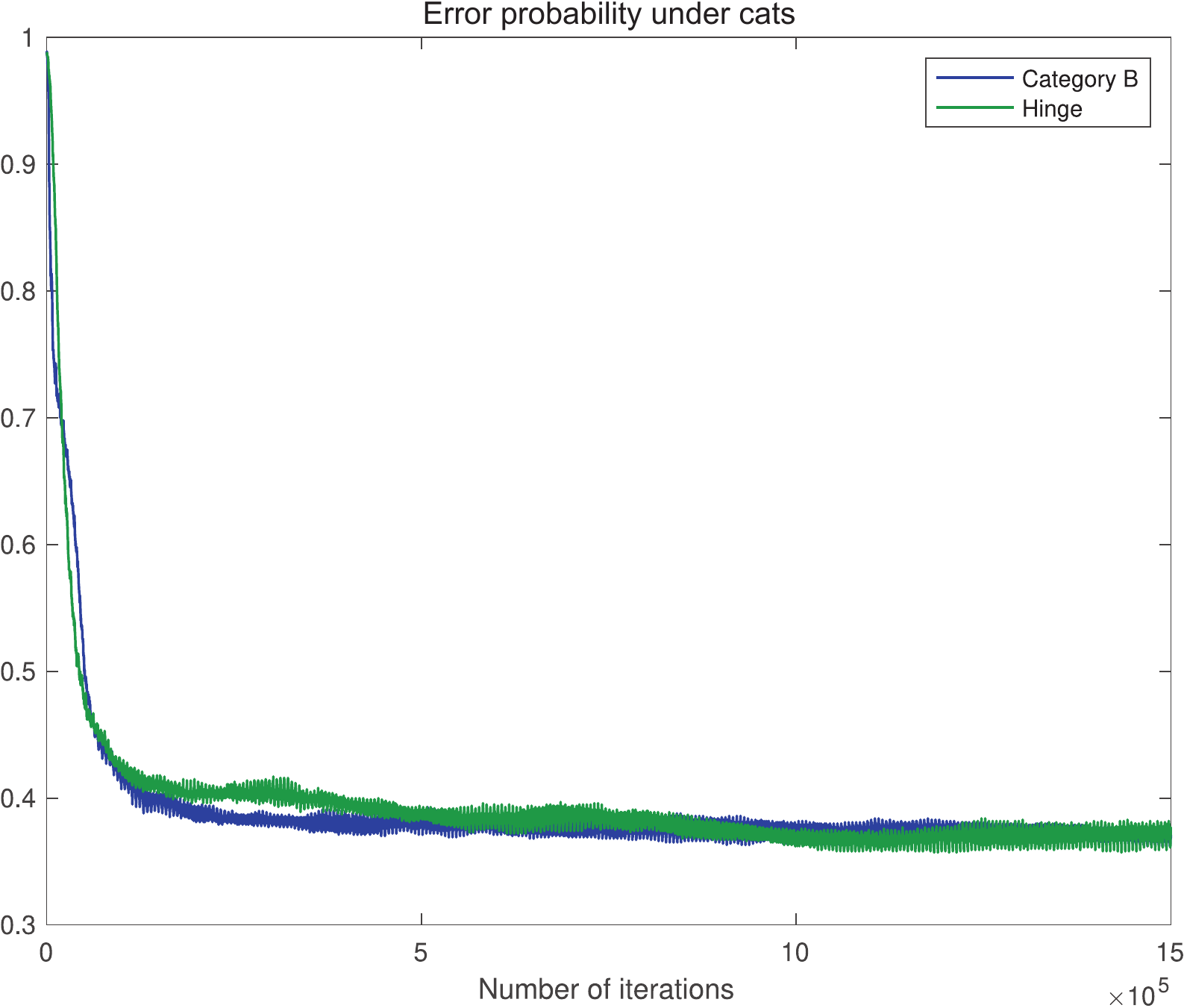}\hfill\includegraphics[width=0.33\hsize]{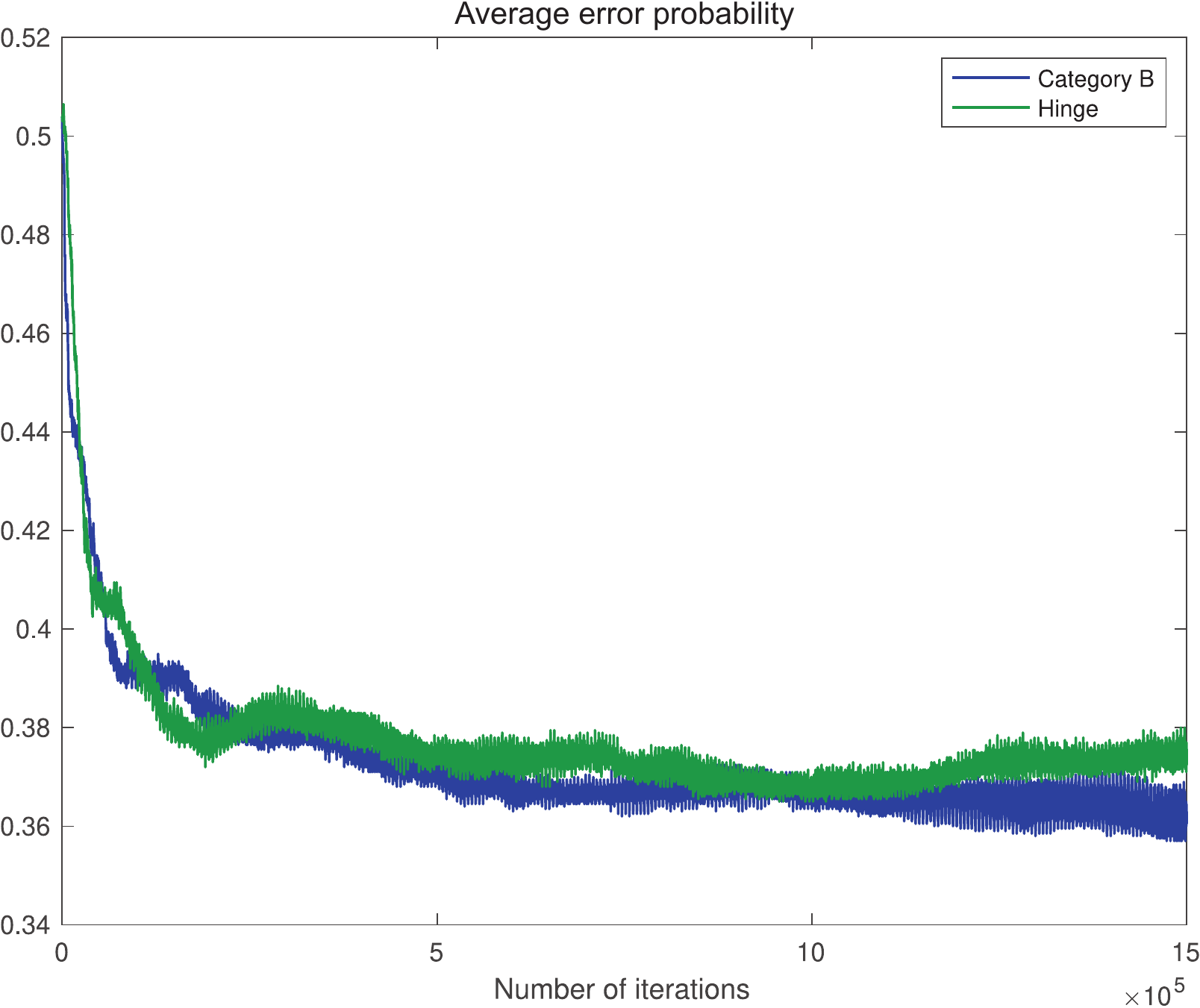}\hfill}
\vskip-0.2cm
\caption{Cats and Dogs}
\vskip0.2cm
\centering
\hbox to \hsize{\hfill\includegraphics[width=0.325\hsize]{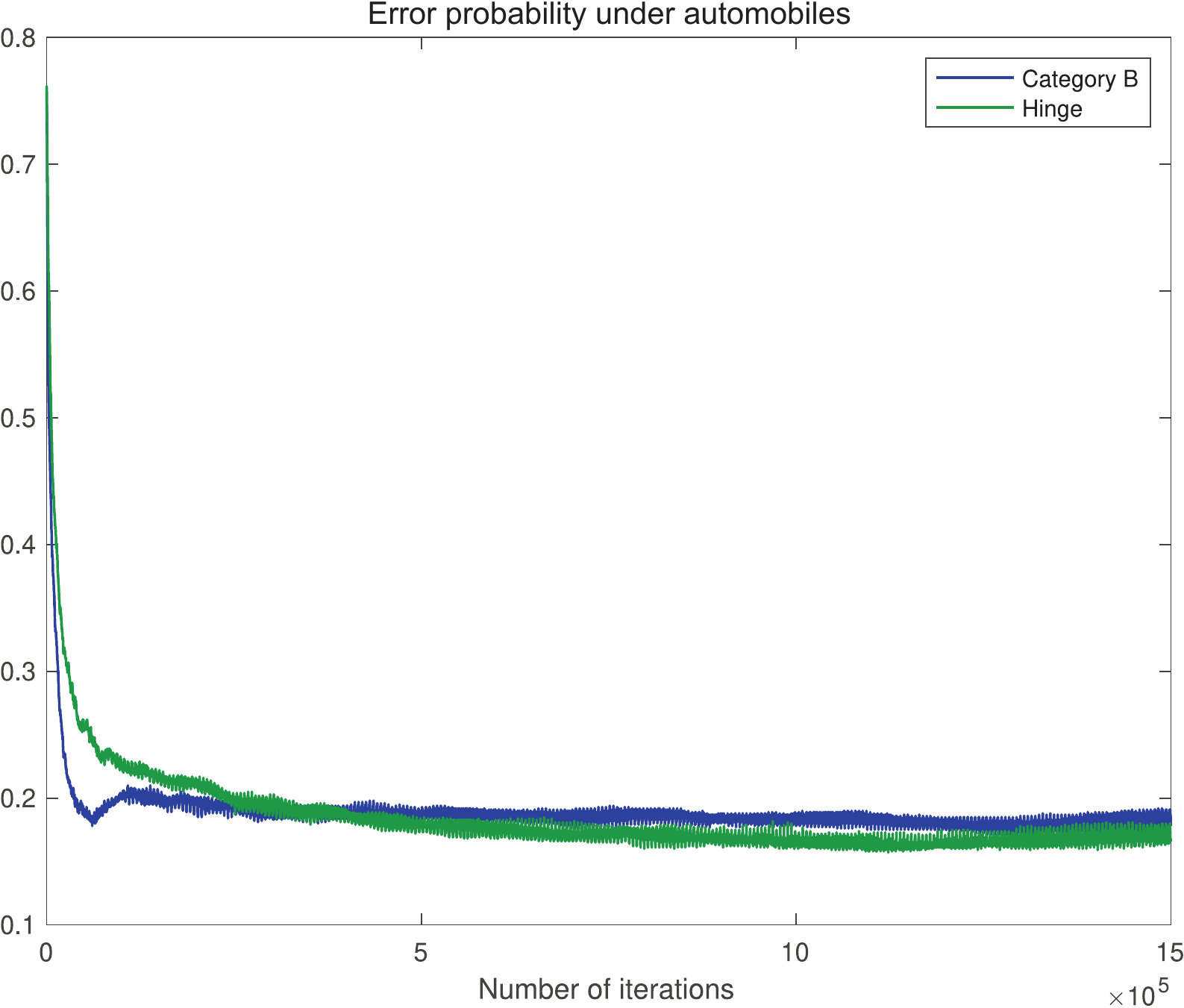}\hfill\includegraphics[width=0.33\hsize]{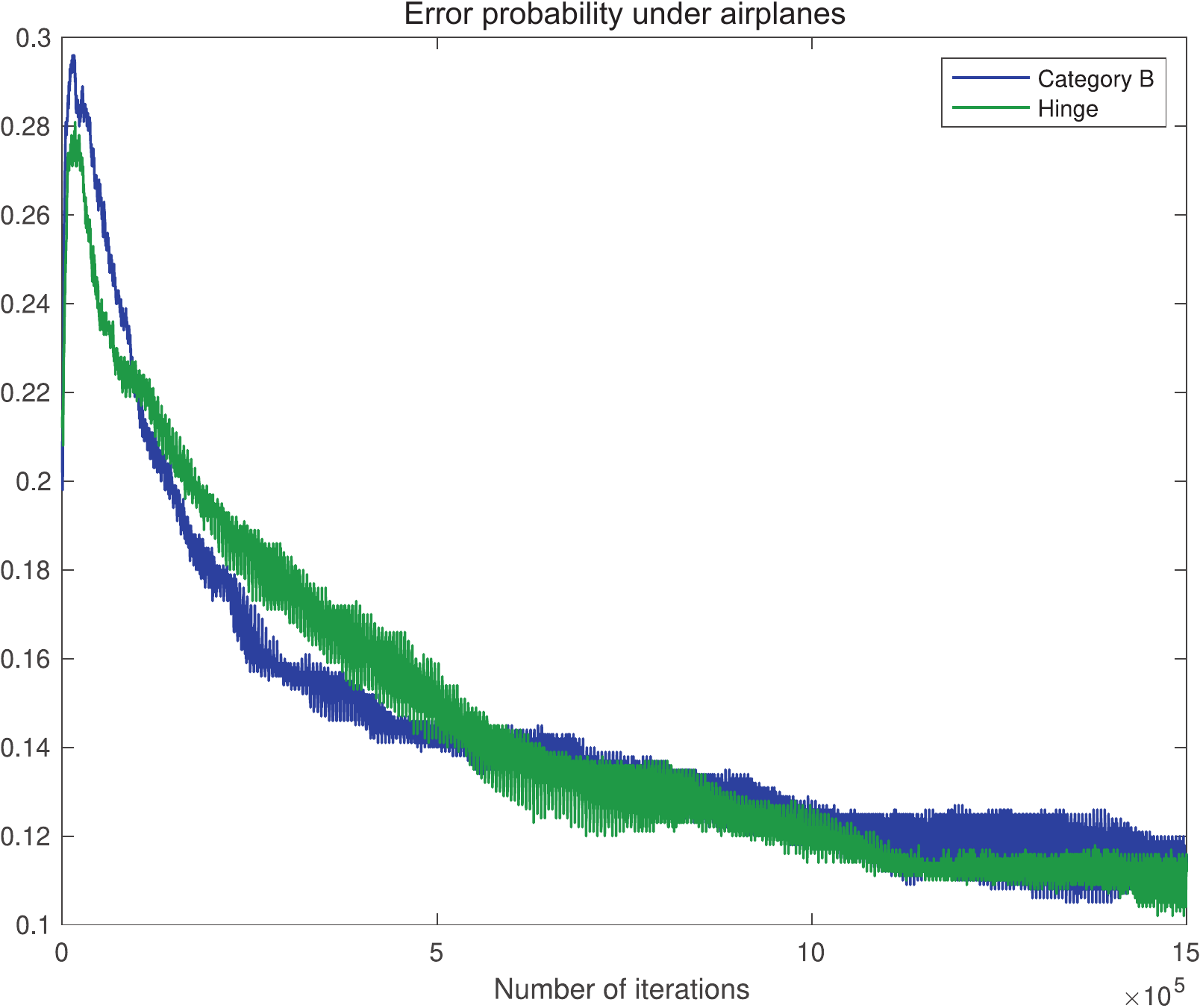}\hfill\includegraphics[width=0.33\hsize]{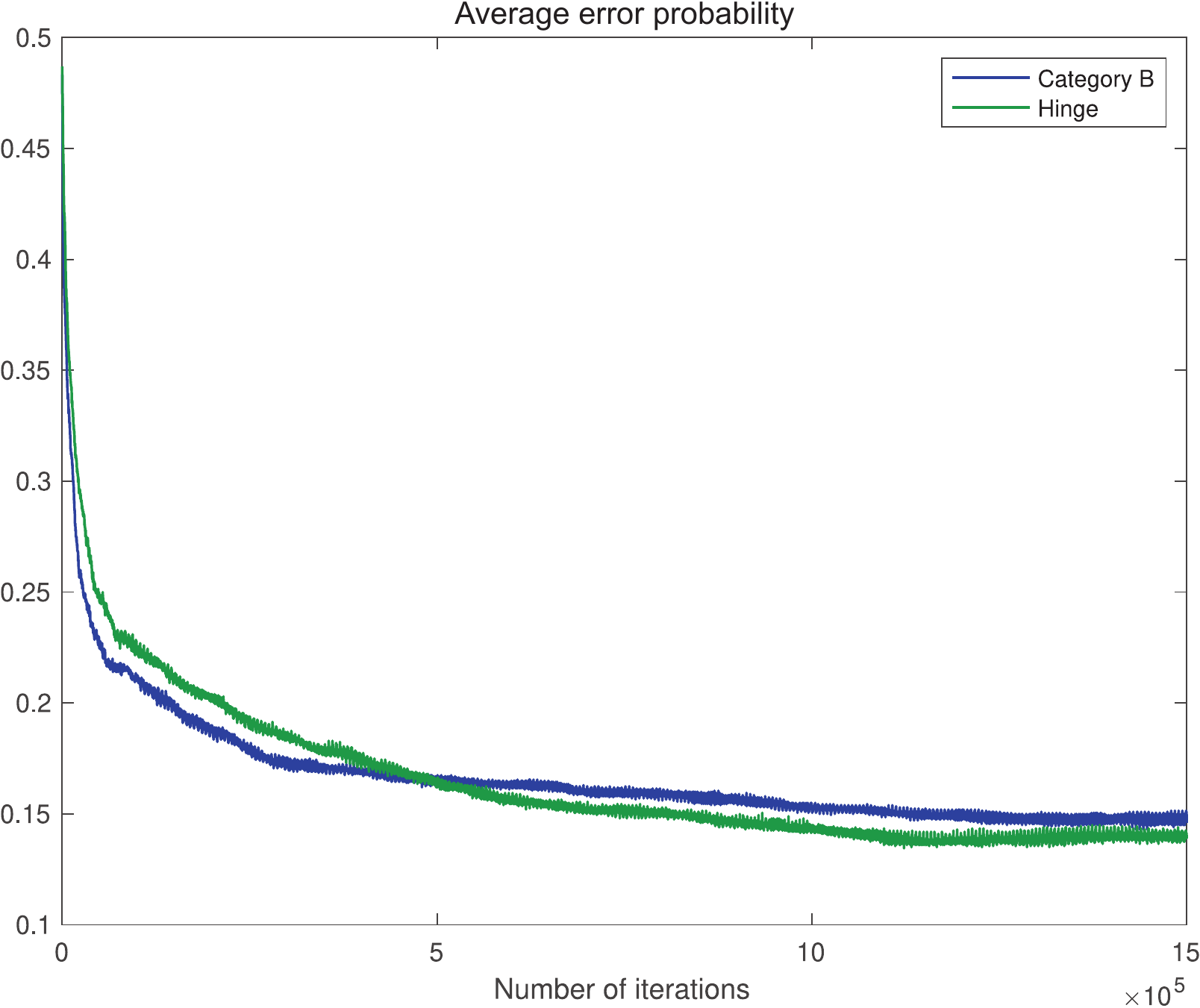}\hfill}
\caption{Automobiles and Airplanes}
\vskip0.2cm
\centering
\hbox to \hsize{\hfill\includegraphics[width=0.33\hsize]{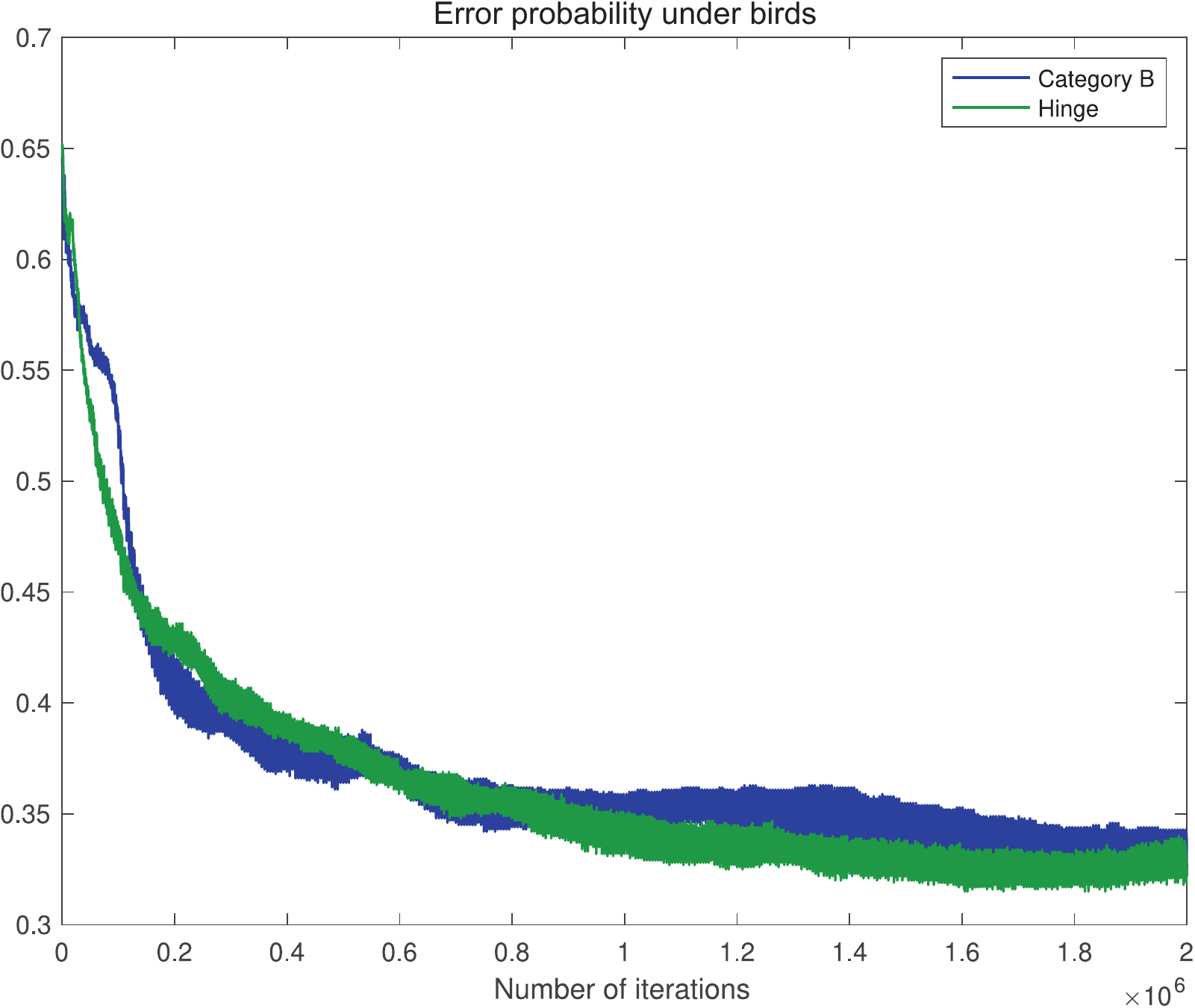}\hfill\includegraphics[width=0.33\hsize]{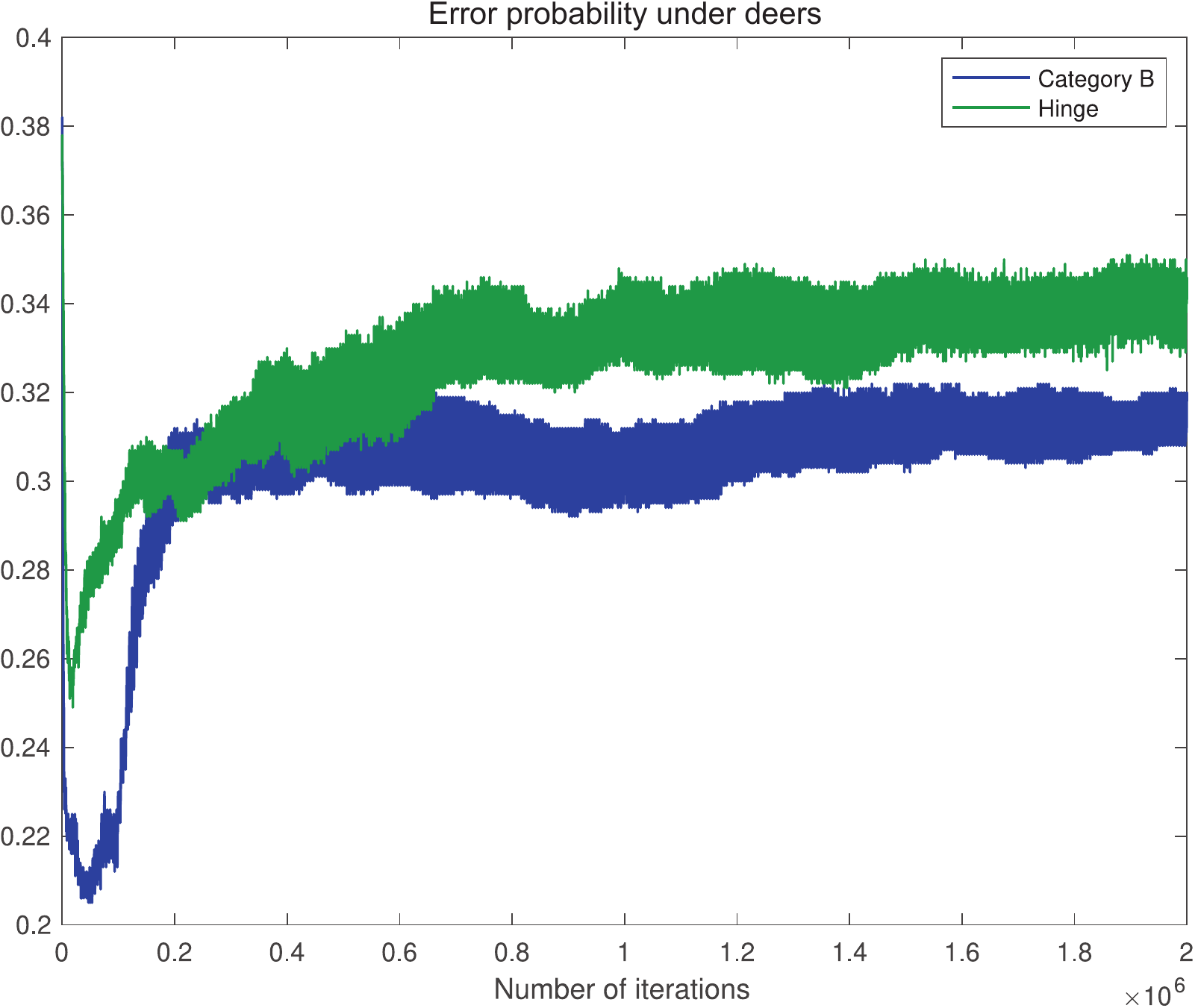}\hfill\includegraphics[width=0.33\hsize]{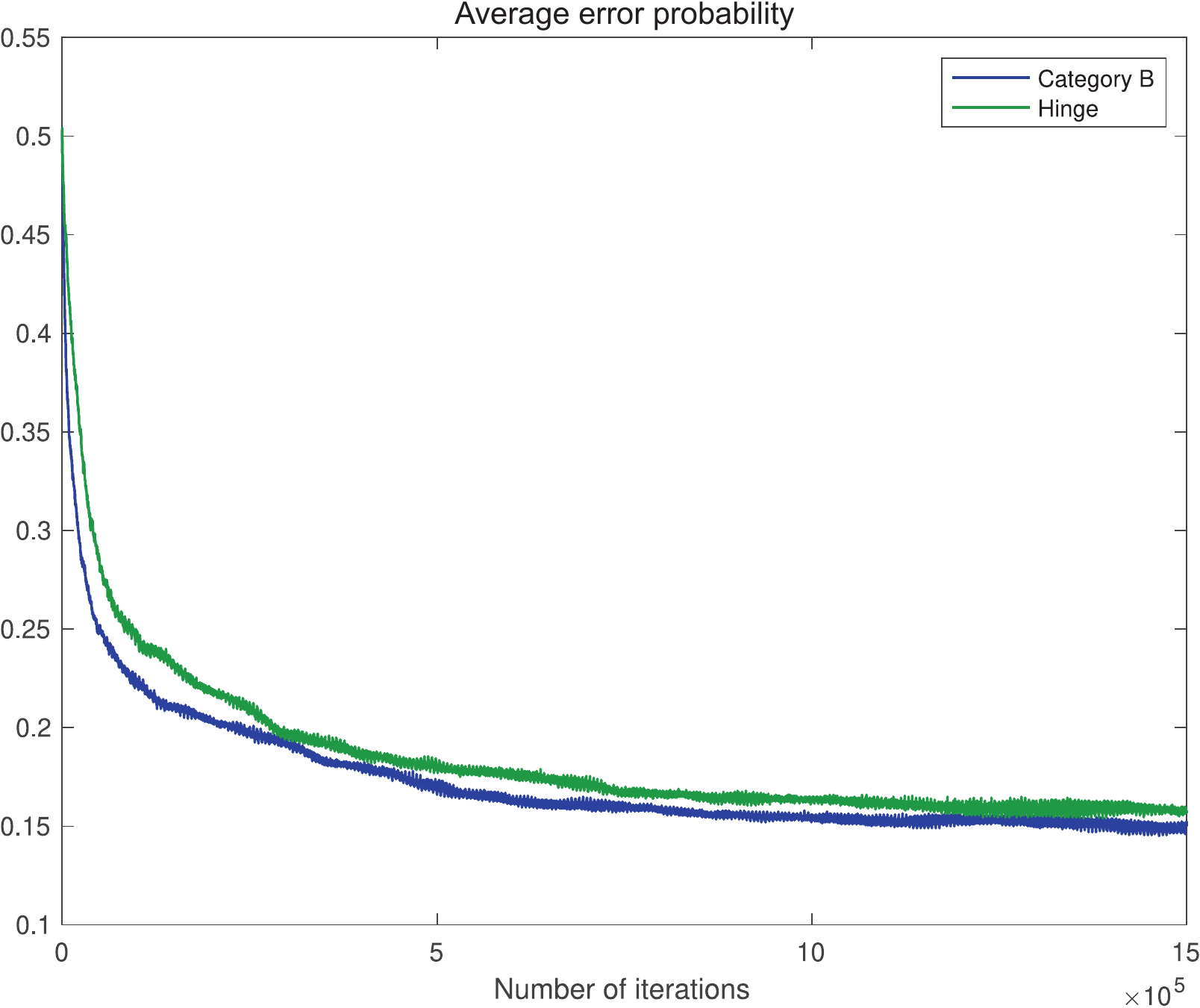}\hfill}
\vskip-0.2cm
\caption{Birds and Deers}
\vskip0.2cm
\centering
\hbox to \hsize{\hfill\includegraphics[width=0.33\hsize]{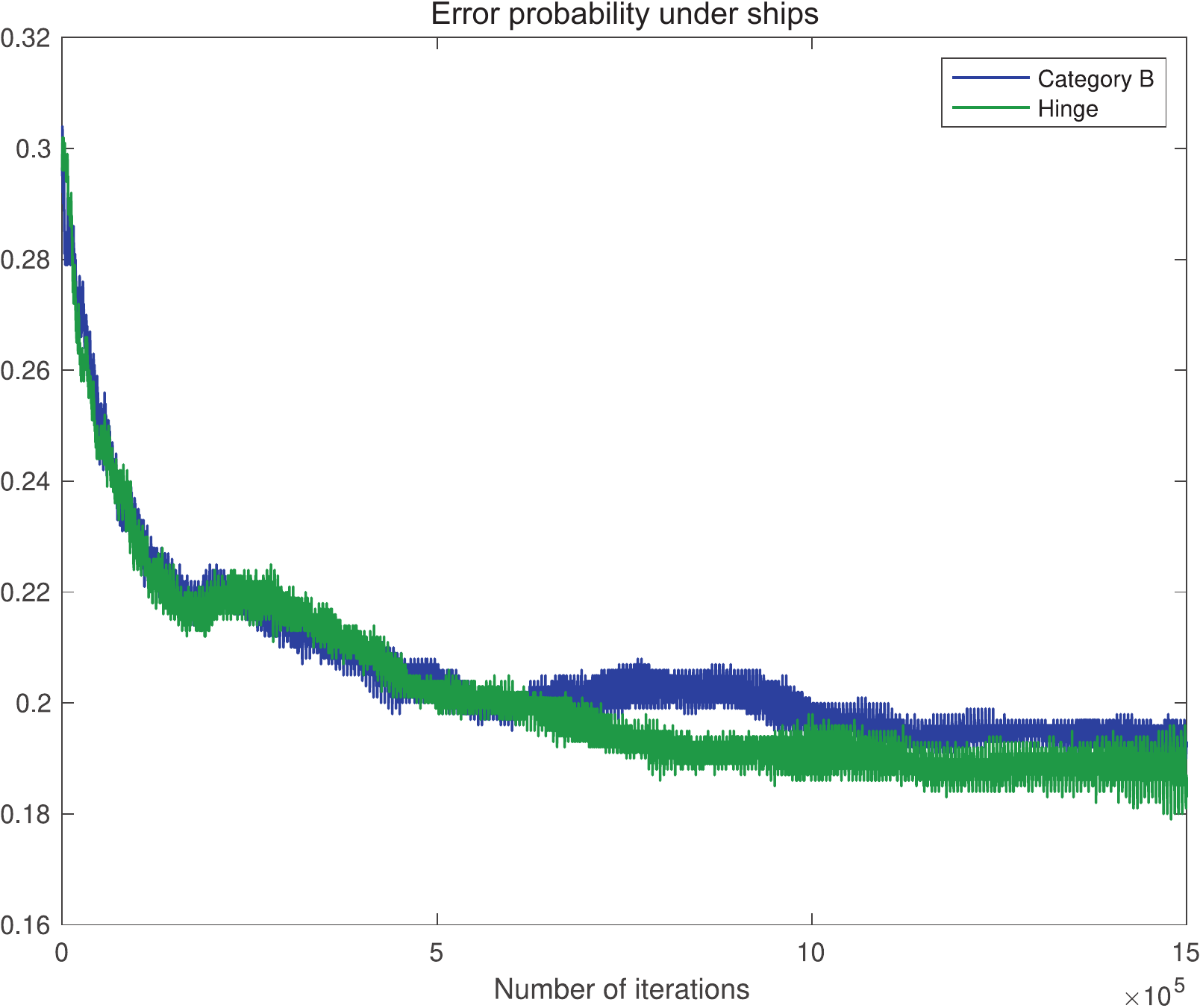}\hfill\includegraphics[width=0.33\hsize]{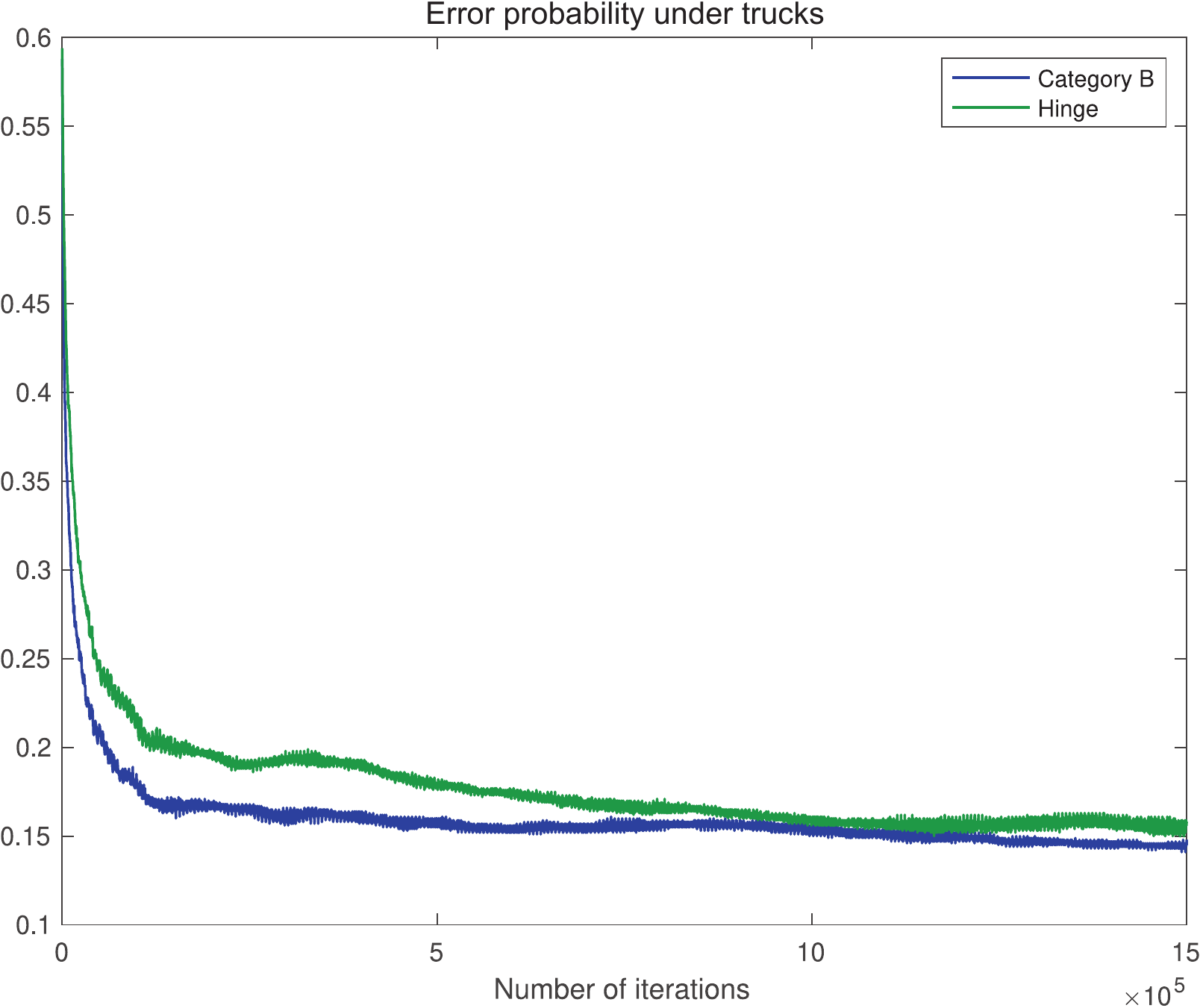}\hfill\includegraphics[width=0.33\hsize]{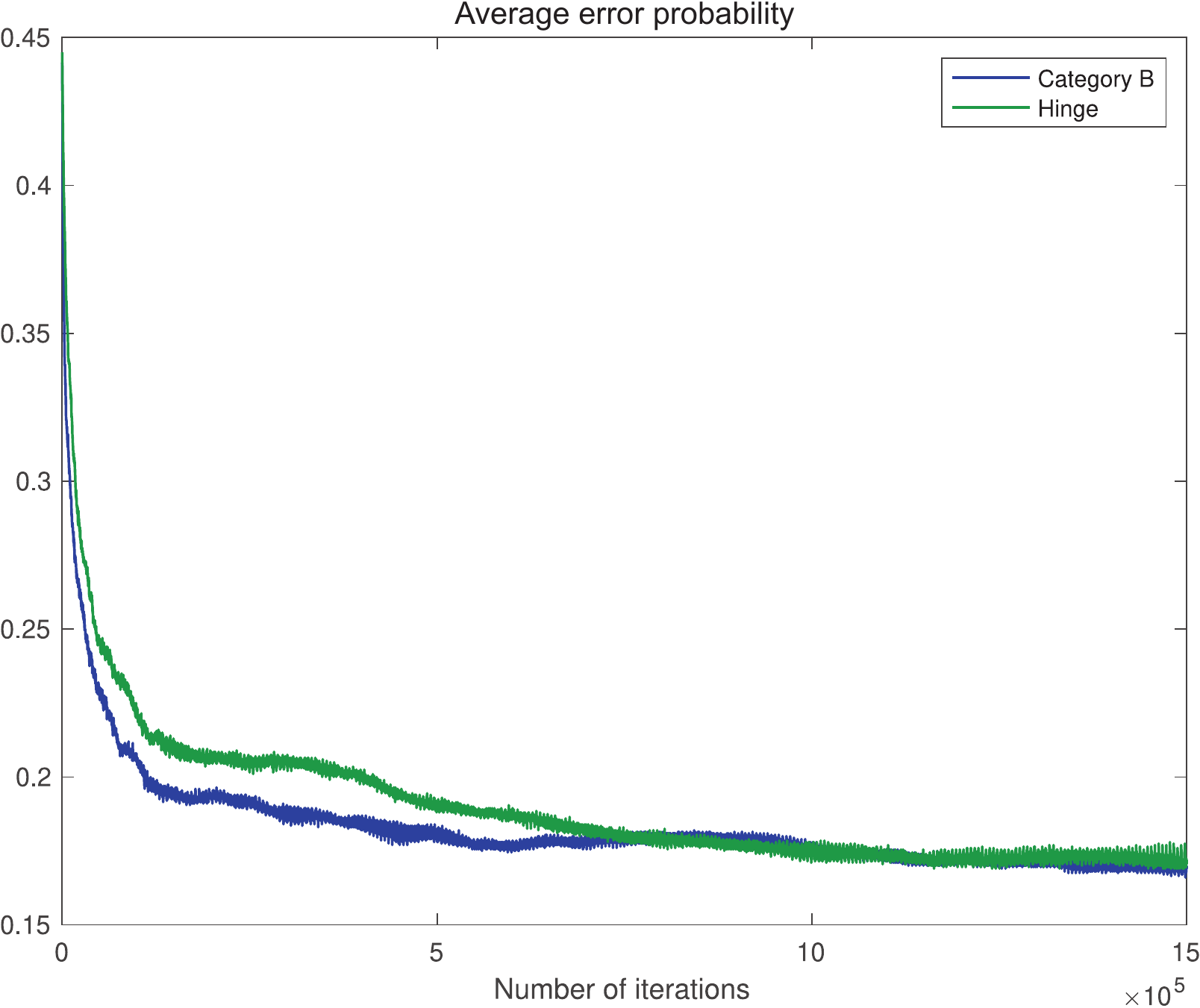}\hfill}
\vskip-0.2cm
\caption{Ships and Trucks}
\vskip0.2cm
\centering
\hbox to \hsize{\hfill\includegraphics[width=0.325\hsize]{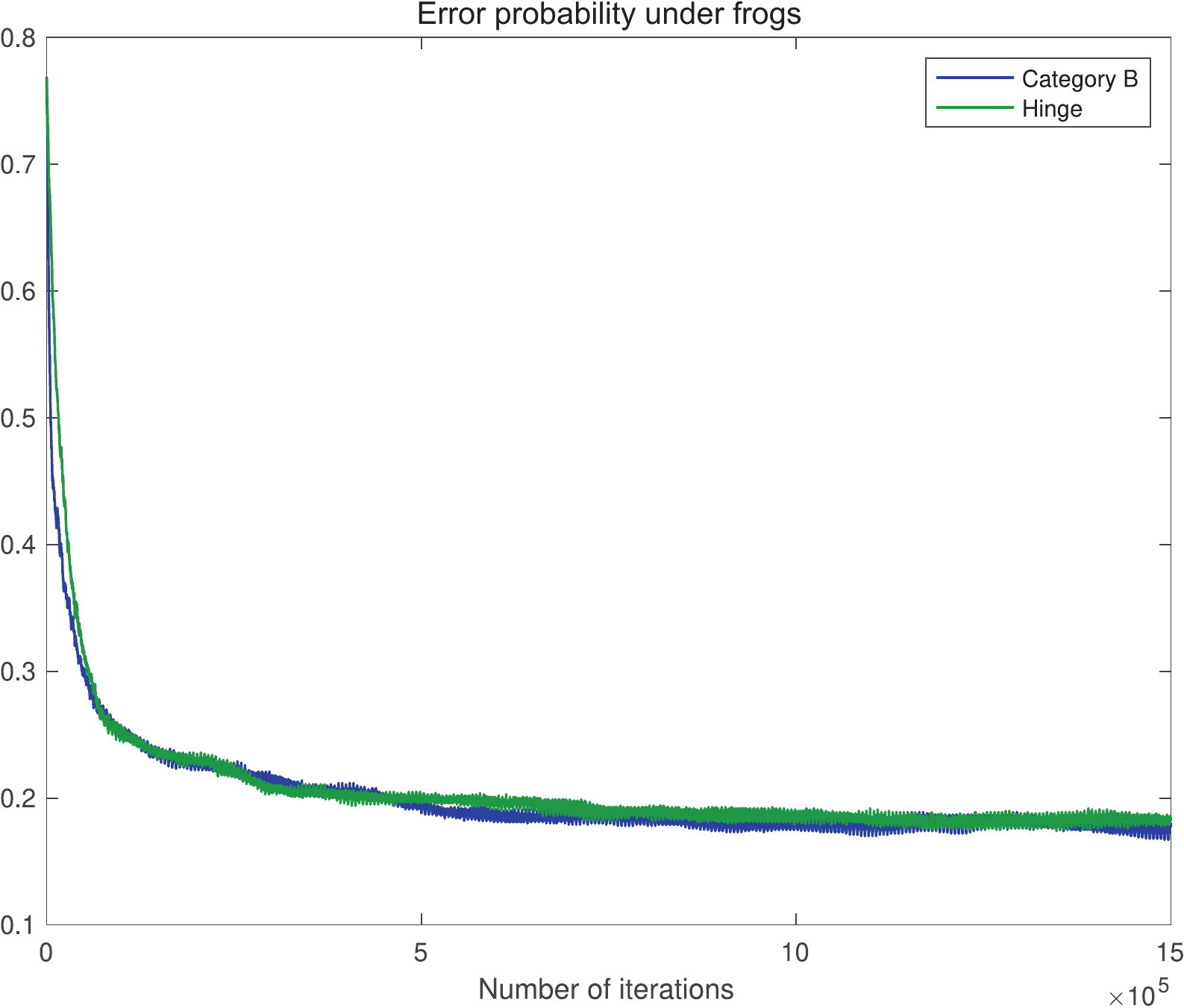}\hfill\includegraphics[width=0.33\hsize]{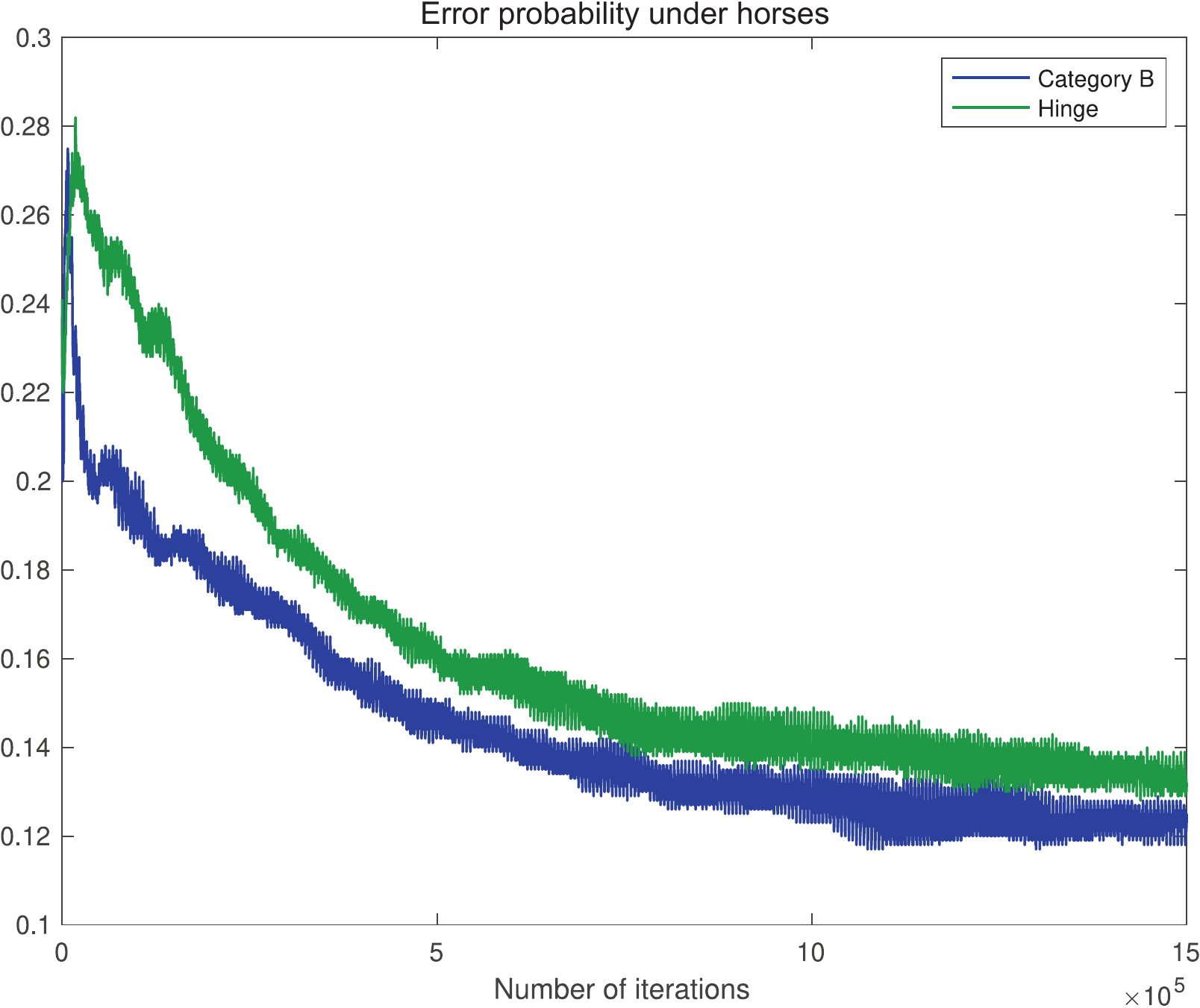}\hfill\includegraphics[width=0.33\hsize]{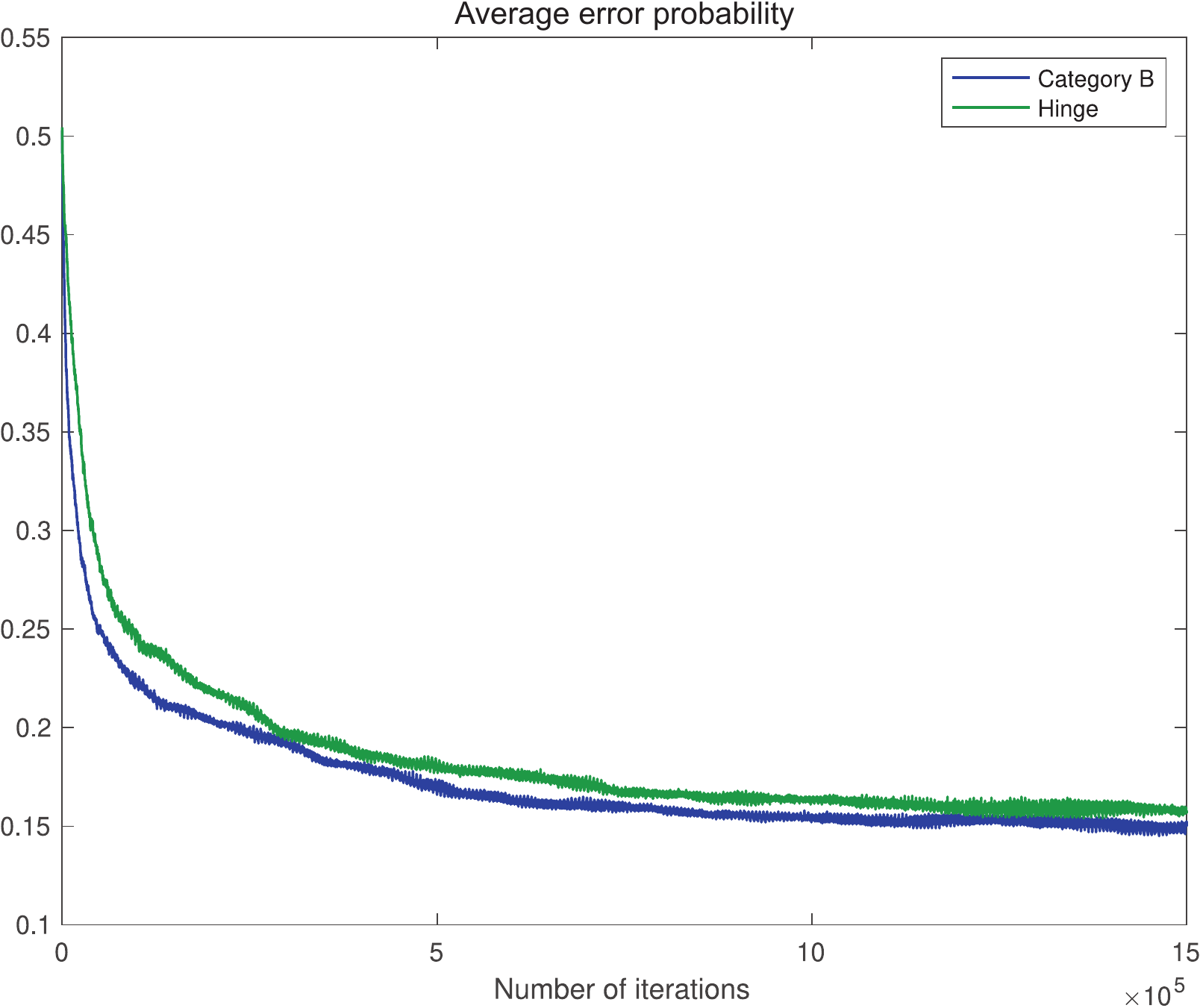}\hfill}
\vskip-0.2cm
\caption{Frogs and Horses}
\end{figure}

CIFAR-10 images were converted from RGB $32\times32\times3$ to grayscale reshaped vectors of size $k=1024$. Furthermore, the mean and the variance of each element of the 1024-length vector was computed from \textit{all} training data and applied to transform each element to make it of zero mean and unit variance. The same transformation was then applied to the testing data. This pre-processing is necessary in order for the adopted network parameter initialization method, proposed in [8], to become appropriate since it assumes that the input is a standard Gaussian.

For the neural network we use two layers with the first being of length $n=100$. The learning rates were selected equal to $\mu=2\times 10^{-5}$ for our Category~B version and $\mu=10^{-5}$ for the Hinge based. In both cases we applied a forgetting factors equal to 0.99. We recycle the training data every time they are exhausted. At each iteration we test the quality of the computed classifiers by applying them to the testing data and computing the resulting error percentage.

We observe that the two competing algorithms have comparable performance. In most of the cases our algorithm exhibits an improvement over the classical scheme and only in one of the cases we observe the opposite. Furthermore, generally speaking, our algorithm exhibits a faster convergence rate, although such claims need far more simulations and, whenever possible, a theoretical analysis, in order to be trusted.
\vspace{-0.2cm}

\section{Conclusions}
\vskip-0.4cm With this work our goal was twofold: First we wanted to understand how existing classifier design techniques are related to each other and, more importantly, to the optimum likelihood ratio test. Second, we wanted to demonstrate that there exists a very simple formulation that can provide an abundance of optimization problems that enjoy the same characteristics as the existing techniques for the classification problem. 
These problems can lend themselves for the development of proper training techniques for neural network based classifiers. The resulting algorithms, compared to existing alternatives, produce classifiers which, in simulations with synthetic data, exhibit a more effective approximation of the optimum likelihood ratio test performance. Additionally, in simulations with real datasets they demonstrate a faster convergence speed attaining, in the limit, most of the times smaller error probabilities. Our immediate future goals include extension of these ideas to the classification of more than two classes and study of the convergence properties of the corresponding training algorithms.
\vspace{-0.25cm}

\section*{Acknowledgment}
\vskip-0.4cm This work was supported by the US National Science Foundation under Grant CIF\,1513373, through Rutgers University.
\vspace{-0.25cm}

\section*{References}
\vskip-0.2cm \small\parskip=0.1cm
[1] Bartlett, P.~L.\ \& Jordan, M.~I.\ \& Mcauliffe, J.~D.\ (2006) Convexity, classification, and risk bounds. {\em Journal of the American Statistical Association, Theory and Methods}, vol. 101, no. 473, pp. 138--156.

[2] Buja, A.\ \& Stuetzle, W.\ \& Shen, Y.\ (2005) Loss functions for binary class probability estimation and classification: Structure and applications. {\em Technical report, University of Pennsylvania}.

[3] Cantrell, C.~D. (2000) {\em Modern Mathematical Methods for Physicists and Engineers}. Cambridge University Press.

[4] Lee, C.-Y.\ \& Xie, S.\ \& Gallagher, P.\ \& Zhang, Z.\ \& Tu, Z.\ (2015) Deeply-supervised nets. In {\em Proceedings of Machine Learning Research}, vol. 38, pp. 562--570.

[5] Choromanska, A.\ \& Henaff, M.\ \& Mathieu, M.\ \& Ben Arous, B.\ \& LeCun, Y.\ (2015) The loss surfaces of multilayer networks. In {\em Proceedings of Machine Learning Research}, vol. 38, pp. 192--204.

[6] Derezi\'nski, M.\ \& Warmuth, M. K.\ (2014). The limits of squared Euclidean distance regularization. {\em Advances in Neural Information Processing Systems}. 4. 2807-2815. 

[7] Eban, E.\ \& Mezuman, E.\ \& Globerson, A. (2014) Discrete Chebyshev classifiers. In {\em Proceedings International Conference on Machine Learning}.

[8] Glorot, X.\ \& Bengio, Y.\ (2010) Understanding the difficulty of training deep feedforward neural networks. In {\em Proceedings International Conference on Artificial Intelligence and Statistics}.

[9] Janocha, K.\ \& Czarnecki, W.~M.\ (2017) On loss functions for deep neural networks in classification. {\em arXiv:1702.05659}. 

[10] Masnadishirazi, H.\ \& Vasconcelos, N. (2009) On the design of loss functions for classification: Theory, robustness to outliers, and SavageBoost. {\em Advances in Neural Information Processing Systems}.

[11] Moulin, P.\ \& Veeravalli, V.~V.\ (2019) {\em Statistical Inference for Engineers and Data Scientists}. Cambridge University Press.

[12] Tang, Y.\ (2013) Deep learning using linear support vector machines. {\em arXiv:1306.0239}.

[13] Tieleman, T.\ \& Hinton, G.\ (2012) Lecture 6.5-rmsprop: Divide the Gradient by a Running Average of Its Recent Magnitude. {\em COURSERA: Neural Networks for Machine Learning}.

\end{document}